\documentclass[journal]{IEEEtran}  
\IEEEoverridecommandlockouts                              

\newif\ifanonymouss
\anonymoussfalse  

\usepackage{array}          
\usepackage{textcomp}       
\usepackage{upgreek}        
\usepackage{stfloats}       
\usepackage{url}            
\usepackage{verbatim}       
\usepackage[normalem]{ulem} 
\useunder{\uline}{\ul}{}
\usepackage[english]{babel} 

\usepackage[space]{cite}    

\makeatletter
\let\NAT@parse\undefined
\makeatother

\usepackage{hyperref}       
\hypersetup{
    colorlinks=true,
    linkcolor=red,
    filecolor=magenta,      
    urlcolor=blue,
    citecolor=blue,
}

\usepackage{graphicx}       

\usepackage[table,xcdraw,dvipsnames]{xcolor} 
\usepackage{colortbl}       

\usepackage{subcaption}     

\usepackage{tabularx}       
\usepackage{threeparttablex}
\usepackage{multirow}       
\usepackage{makecell}       
\usepackage{booktabs}       

\usepackage{amsmath}        
\usepackage{amsfonts}       
\usepackage{amssymb}        
\usepackage{amsthm}       

\theoremstyle{definition}

\theoremstyle{remark}

\usepackage[ruled,vlined,linesnumbered]{algorithm2e} 
\usepackage{algorithmic}    

\usepackage{CJKutf8}

\title{\LARGE \bf
NRGS-SLAM: Monocular Non-Rigid SLAM for Endoscopy via Deformation-Aware 3D Gaussian Splatting
}

\ifanonymouss
\else
\author{Jiwei Shan, Zeyu Cai, Yirui Li, Yongbo Chen, Lijun Han, Yun-hui Liu, Hesheng Wang$^{\dag}$ and Shing Shin Cheng$^{\dag}$          
\thanks{Jiwei Shan, Zeyu Cai, Yirui Li, Shing Shin Cheng and Yun-hui Liu are with the Department of Mechanical and Automation Engineering and T Stone Robotics Institute, The Chinese University of Hong Kong, Hong Kong. Yongbo Chen, Lijun Han and Hesheng Wang are with the School of Automation and Intelligent Sensing, the State Key Laboratory of Avionics Integration and Aviation System of-Systems Synthesis, and Shanghai Key Laboratory of Navigation and Location Based Services, Shanghai Jiao Tong University, Shanghai 200240, China }
\thanks{$^{\dag}$Corresponding author: Shing Shin Cheng and Hesheng Wang}
}
\fi

\begin{document}
\maketitle
\thispagestyle{empty}
\pagestyle{empty}

\begin{abstract}
Visual simultaneous localization and mapping (V-SLAM) is a fundamental capability for autonomous perception and navigation. However, endoscopic scenes violate the rigidity assumption due to persistent soft-tissue deformations, creating a strong coupling ambiguity between camera ego-motion and intrinsic deformation. Although recent monocular non-rigid SLAM methods have made notable progress, they often lack effective decoupling mechanisms and rely on sparse or low-fidelity scene representations, which leads to tracking drift and limited reconstruction quality. To address these limitations, we propose NRGS-SLAM, a monocular non-rigid SLAM system for endoscopy based on 3D Gaussian Splatting. To resolve the coupling ambiguity, we introduce a deformation-aware 3D Gaussian map that augments each Gaussian primitive with a learnable deformation probability, optimized via a Bayesian self-supervision strategy without requiring external non-rigidity labels. Building on this representation, we design a deformable tracking module that performs robust coarse-to-fine pose estimation by prioritizing low-deformation regions, followed by efficient per-frame deformation updates. A carefully designed deformable mapping module progressively expands and refines the map, balancing representational capacity and computational efficiency. In addition, a unified robust geometric loss incorporates external geometric priors to mitigate the inherent ill-posedness of monocular non-rigid SLAM.
Extensive experiments on multiple public endoscopic datasets demonstrate that NRGS-SLAM achieves more accurate camera pose estimation (up to 50\% reduction in RMSE) and higher-quality photo-realistic reconstructions than state-of-the-art methods. Comprehensive ablation studies further validate the effectiveness of our key design choices. Source code will be publicly available upon paper acceptance.
\end{abstract}

\section{Introduction}
Visual Simultaneous Localization and Mapping (V-SLAM) is a fundamental capability enabling autonomous systems to navigate the physical world. It allows for the online estimation of camera motion while concurrently reconstructing unknown surroundings, relying on the fundamental assumption of environmental rigidity \cite{cadena2017past}. Over the past decades, substantial progress has been made in both academia and industry \cite{cadena2017past}. Methodologies have evolved from early probabilistic and geometric formulations based on extended Kalman filtering \cite{davison2007monoslam}, feature correspondences \cite{mur2015orb,mur2017orb}, and direct pixel alignment \cite{engel2017direct}, to recent approaches leveraging differentiable rendering for dense, photo-realistic scene representation \cite{sucar2021imap,matsuki2024gaussian,keetha2024splatam,yan2024gs,zhang2025hi,li2025pg}.

\begin{figure}[t]
\centerline{\includegraphics[width=0.5\textwidth]{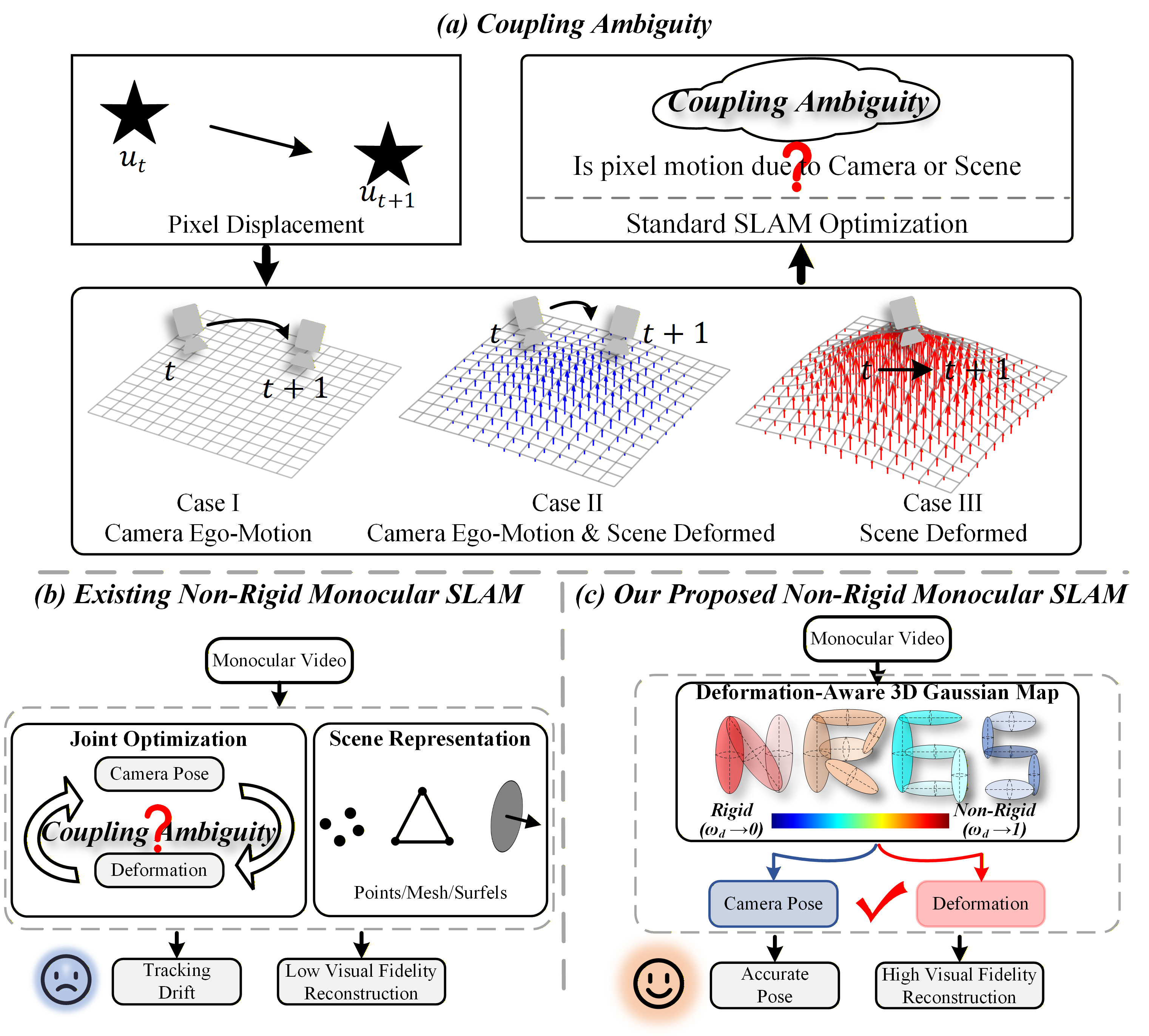}}
\caption{(a) Illustration of the Coupling Ambiguity: In monocular non-rigid scenarios, pixel displacement arise from a complex combination of camera ego-motion and intrinsic tissue deformation. This creates a fundamental coupling ambiguity, making it difficult to distinguish rigid motion from non-rigid dynamics or their combination. (b) Existing methods often struggle to effectively decouple these factors during joint optimization and rely on sparse representations (e.g., points or meshes), resulting in tracking drift and low visual fidelity. (c) Our proposed NRGS-SLAM introduces a deformation-aware 3D Gaussian map. By learning a per-primitive deformation probability (visualized by the color spectrum ranging from blue (rigid) to red (non-rigid)), our system explicitly decouples camera tracking from deformation updates, enabling both accurate pose estimation and high-fidelity photo-realistic reconstruction.}
    \label{fig:fig1}
\end{figure}

However, in endoscopic surgical scenarios, the assumptions underlying conventional V-SLAM formulations are violated~\cite{schmidt2024tracking}. 
In particular, environmental rigidity no longer holds, as organ and soft-tissue surfaces undergo persistent non-rigid deformations driven by physiological motion or interactions with surgical instruments. 
While existing dynamic V-SLAM systems can handle moving objects in natural scenes~\cite{yu2018ds,li2025pg,xu2024dg,li20254d}, they typically rely on clear semantic distinctions between moving foreground objects and a static background. 
In contrast, endoscopic scenes exhibit complex and continuous deformations in which rigid and non-rigid regions are not semantically separable. 
Consequently, as illustrated in Fig.~\ref{fig:fig1}(a), observed pixel variations may arise from camera ego-motion, intrinsic scene deformation, or a combination of both. This {coupling ambiguity} compromises the geometric constraints assumed by V-SLAM, posing significant challenges for accurate tracking and reconstruction.

To address these challenges, monocular deformable SLAM has gained increasing attention in recent years. Existing methods typically rely on traditional scene representations, such as meshes~\cite{lamarca2020defslam,gomez2021sd}, surfels~\cite{lamarca2022direct}, or sparse point clouds~\cite{rodriguez2024nr,rodriguez2022tracking}, and model deformation using techniques such as Shape-from-Template(SfT)~\cite{lamarca2018camera}, Non-Rigid Structure-from-Motion (NRSfM) \cite{parashar2017isometric}, or embedded deformation graphs \cite{rodriguez2024nr}. Despite achieving promising results, significant challenges remain.  As shown in Fig.~\ref{fig:fig1}(b), a critical limitation of these approaches is the lack of explicit mechanisms to effectively reduce the coupling between camera ego-motion and intrinsic scene deformation. Without such decoupling, the optimization process often misattributes visual changes, leading to tracking drift. Moreover, the scene representations used in these methods are often sparse or lack texture detail, limiting the visual fidelity of the reconstruction. Although recent advances in differentiable rendering have shown strong potential for dense and photo-realistic modeling and have been successfully integrated into rigid V-SLAM systems~\cite{sucar2021imap,matsuki2024gaussian,keetha2024splatam,yan2024gs,zhang2025hi,li2025pg}, their extension to non-rigid endoscopic SLAM remains underexplored.

In this paper, we propose {NRGS-SLAM}, a monocular deformable SLAM system built upon recent differentiable rendering technology, specifically 3D Gaussian Splatting (3DGS)~\cite{kerbl20233d}. 
As shown in Fig.~\ref{fig:overview}, our system comprises four main components: (i) a deformation-aware 3D Gaussian map, (ii) deformable tracking, (iii) deformable mapping, and (iv) measurement pre-processing.
At the core of NRGS-SLAM is the \textit{deformation-aware 3D Gaussian map}. Following recent advances in surgical scene reconstruction~\cite{wu20244d,yang2024deformable,long2025surgical,shan2025deformable,yang2024deform3dgs}, we represent the scene using 3D Gaussian primitives in a canonical space, coupled with a continuous deformation field. To address the challenge of {coupling ambiguity}, our core idea is to explicitly model the intrinsic tendency of each scene primitive to undergo non-rigid motion, thereby distinguishing between low-deformation and high-deformation regions to ensure they fulfill distinct roles. To this end, as shown in Fig.~\ref{fig:fig1}(c), we augment each 3D Gaussian primitive with a {deformation probability}, treating it as an explicit, learnable property of the scene representation. Naturally integrated into the differentiable Gaussian rasterization pipeline, this attribute generates a dense, pixel-aligned {deformation confidence} map that adaptively modulates the contribution of image pixels during optimization. Given the lack of ground-truth non-rigidity labels, we propose a Bayesian self-supervision strategy. This strategy estimates a posterior deformation likelihood from image observations, serving as a stable pseudo-supervisory signal. This design eliminates the need for external labels while enabling the deformation-aware map to progressively calibrate itself during the online reconstruction process.

Building on this foundation, we carefully design the \textit{deformable tracking module}. We employ a robust coarse-to-fine camera pose estimation strategy. 
Leveraging the learned deformation probabilities, the tracking process prioritizes regions predicted to be reliable (i.e., having low deformation probability). 
This explicitly mitigates the coupling between camera motion and scene deformation, ensuring robust pose estimation. 
Once a reliable pose is established, we perform an efficient per-frame deformation update to capture instantaneous scene changes. In the \textit{deformable mapping} module, the deformation-aware 3D Gaussian map is progressively expanded to incorporate new observations as the camera explores previously unseen regions. 
A global bundle deformable adjustment is then performed to jointly refine the map and keyframe poses. 
During this process, we generate the supervisory signals for learning deformation probabilities and employ an adaptive mechanism to regulate the deformation field, thereby striking a balance between representational fidelity and computational efficiency. Finally, to alleviate the inherent ill-posedness of jointly recovering camera poses and scene deformation from monocular data, the \textit{measurement pre-processing} module incorporates geometric priors derived from large-scale foundation model. We introduce a unified geometric loss within a robust optimization framework to integrate these priors, which enables the system to benefit from geometric guidance while remaining resilient to the prediction noise and domain shifts commonly encountered in endoscopic imagery.

Overall, our main contributions are summarized as follows:
\begin{itemize}
    \item We introduce a {deformation-aware 3D Gaussian map} augmented with a learnable deformation probability. This attribute provides an explicit mechanism to reduce the coupling between camera ego-motion and intrinsic scene deformation and is supervised through a Bayesian self-supervision strategy.
    \item We propose a deformable tracking module, which leverages deformation probabilities for coarse-to-fine pose estimation and efficient per-frame deformation updates.
    \item We propose a deformable mapping module integrating progressive map extension, adaptive deformation field management, and global deformable bundle adjustment.
    \item We propose a unified robust geometric loss that integrates external geometric priors, effectively mitigating the ill-posed nature of non-rigid SLAM while maintaining robustness to prediction noise.
\end{itemize}
Extensive experiments on multiple public endoscopic datasets demonstrate that NRGS-SLAM achieves more accurate camera pose estimation and higher-quality deformable endoscopic scene reconstruction than state-of-the-art methods.
\section{Related Work}
\label{sec:related work}
In this work, we focus on monocular non-rigid SLAM based on 3D Gaussian Splatting for deformable endoscopic scenes.
Accordingly, this section reviews the literature most relevant to our framework across two primary domains:
1) monocular non-rigid SLAM, with a particular emphasis on endoscopic environments; and
2) Gaussian Splatting-based SLAM.

\subsection{Monocular Non-rigid SLAM}
While numerous state-of-the-art V-SLAM systems have been adapted to the endoscopic domain to mitigate challenges such as texture scarcity and illumination variations, they fundamentally retain the assumption of environmental rigidity. 
Therefore, we omit a detailed review here and refer readers to~\cite{schmidt2024tracking}.
In contrast, non-rigid SLAM relaxes the rigidity assumption, allowing scene geometry to deform over time. 
This introduces significant complexity, as both the camera pose and the 3D structure become time-varying unknowns, rendering the problem highly ill-posed.
DefSLAM~\cite{lamarca2020defslam} pioneered the explicit modeling of such non-rigid deformations within a monocular framework. 
It utilizes a triangular mesh for scene representation, employing Shape-from-Template (SfT)~\cite{lamarca2018camera} for frame-rate deformation and camera pose estimation. 
Simultaneously, the mapping thread updates the template at the keyframe rate via Non-Rigid Structure-from-Motion (NRSfM)~\cite{parashar2017isometric}.
Building on this, SD-DefSLAM~\cite{gomez2021sd} incorporates Lucas-Kanade (LK) optical flow to enhance data association robustness and integrates an ORB bag-of-words model to support relocalization under mild deformations.
However, the topological constraints inherent to triangular meshes limit their ability to reconstruct surfaces with discontinuities or holes~\cite{rodriguez2024nr}.
To address these representational limitations, DSDT~\cite{lamarca2022direct} proposes a monocular approach (requiring stereo initialization) that estimates camera pose and deformation by independently tracking deformable surfel projections.
Alternatively, \cite{rodriguez2022tracking} models the scene using sparse point clouds, constraining geometry with As-Rigid-As-Possible (ARAP) deformation and Elasticity Description (ED) models.
Subsequently, NR-SLAM~\cite{rodriguez2024nr} integrates a Dynamic Deformation Graph (DDG) into the framework of~\cite{rodriguez2022tracking} to develop a complete monocular non-rigid SLAM system. 

Nevertheless, as illustrated in Fig.~\ref{fig:fig1}(b), these approaches generally lack explicit mechanisms to decouple camera motion from scene dynamics and often fail to capture high-frequency tissue details.
In this work, based on 3D Gaussian Splatting, we propose a deformation-aware scene modeling and optimization framework that effectively alleviates the coupling between camera ego-motion and scene deformation, while preserving high-fidelity, photo-realistic reconstruction from monocular endoscopic observations.

\subsection{Gaussian Splatting-based SLAM}

3D Gaussian Splatting (3DGS)~\cite{kerbl20233d} has emerged as a powerful scene representation, offering high-fidelity, real-time novel view synthesis by modeling scenes via anisotropic 3D Gaussians. 
Leveraging these capabilities, numerous concurrent works have integrated 3DGS into dense SLAM frameworks~\cite{matsuki2024gaussian,keetha2024splatam,yan2024gs,zhang2025hi,tosi2024nerfs}. 
These approaches typically synthesize RGB and depth images from the 3D Gaussian map, facilitating joint optimization of scene parameters and camera poses by minimizing photometric and geometric residuals. 
However, these methods are primarily tailored for static indoor or outdoor environments. Some subsequent GS-based SLAM~\cite{li2025pg,xu2024dg,li20254d} have improved robustness in general dynamic scenes by using semantic segmentation to filter or explicitly model dynamic objects. 
However, such semantic-based strategies prove ineffective in endoscopy, where rigid and non-rigid regions are not semantically separable.
Alternatively, 4DTAM~\cite{matsuki20254dtam} attempts to model dynamics by coupling a canonical space with a Multi-Layer Perceptron deformation field. 
Nevertheless, this approach relies heavily on dense depth input, precluding its application in monocular endoscopic scenarios. 
Furthermore, it fails to effectively decouple camera ego-motion from scene deformation, leaving the fundamental ambiguity between tracking and non-rigid motion unresolved.

More recently, several works have adapted GS-based SLAM to endoscopic scenes. 
EndoGSLAM~\cite{wang2024endogslam} extends the SplatAM framework~\cite{keetha2024splatam} with endoscopy-specific adaptations, showing promising results in static environments. 
Endo-2DTAM~\cite{11128637} utilizes 2D Gaussian Splatting~\cite{huang20242d} to enhance surface reconstruction accuracy. 
EndoFlow-SLAM~\cite{wu2025endoflow} incorporates optical flow losses as geometric constraints to regularize both structure and motion estimation. 
Despite these advances, existing endoscopic GS-based SLAM methods face two critical limitations. 
First, they predominantly assume a static scene, rendering them ill-equipped to handle deformations. 
Second, they typically rely on dense depth input, which conflicts with the standard monocular setup of clinical endoscopes. Notably, prior to these developments, ENeRF-SLAM~\cite{shan2024enerf} and DDS-SLAM~\cite{shan2024dds} achieved high-fidelity reconstruction using differentiable rendering techniques. 
However, the substantial computational overhead associated with implicit neural representations limits their real-time applicability.

In this work, we propose a monocular non-rigid SLAM framework for endoscopy based on deformation-aware 3D Gaussian Splatting. 
As depicted in Fig.~\ref{fig:fig1}(c), our method effectively alleviates the coupling between camera motion and scene deformation, enabling accurate camera pose estimation and photo-realistic scene reconstruction.
\section{Problem Formulation}
\label{sec:3dgs}
\subsection{Preliminary}
In this section, we first introduce the preliminary knowledge of 3D Gaussian Splatting (3DGS) and deformable 3DGS. Subsequently, we present an overview of our SLAM system.
\subsubsection{3D Gaussian Splatting}\label{subsec:3dgs}
3DGS~\cite{kerbl20233d} is a real-time differentiable rendering technique that models scenes using a set of $N$ anisotropic Gaussian primitives 
$\mathcal{G} = \left\{ \mathbf{G}_i = \left( \boldsymbol{\mu}_i, \mathbf{s}_i, \mathbf{q}_i, \sigma_i, \mathbf{c}_i \right) \right\}_{i=1}^N$. Each Gaussian primitive $\mathbf{G_i}$ influences a spatial point $\mathbf{x} \in \mathbb{R}^3$ according to:
\begin{equation}
    \mathbf{G_i}(\mathbf{x}) = \exp \left( -\frac{1}{2} (\mathbf{x} - \boldsymbol{\mu}_i)^\top \boldsymbol{\Sigma}_i^{-1} (\mathbf{x} - \boldsymbol{\mu}_i) \right),
\end{equation}
where $\boldsymbol{\mu}_i$ denotes the center and $\boldsymbol{\Sigma}_i$ is the covariance matrix. To ensure positive semi-definiteness, $\boldsymbol{\Sigma}_i$ is decomposed into a scaling matrix $\mathbf{S}_i$ and a rotation matrix $\mathbf{R}_i$ as $\boldsymbol{\Sigma}_i = \mathbf{R}_i \mathbf{S}_i \mathbf{S}_i^\top \mathbf{R}_i^\top$, where $\mathbf{S}_i$ and $\mathbf{R}_i$ are constructed from a 3D scaling vector $\mathbf{s}_i$ and a unit quaternion $\mathbf{q}_i$, respectively. Each Gaussian also carries a learnable opacity scalar $\sigma_i \in [0, 1]$ and view-dependent color features represented by spherical harmonics coefficients $\mathbf{c}_i$.

During rendering, the 3D Gaussian primitives are projected onto the 2D image plane under the current camera pose $\mathbf{T}$. 
The projected 2D covariance matrix $\boldsymbol{\Sigma}'$ and center $\boldsymbol{\mu}'$ in camera coordinates are computed as $\boldsymbol{\Sigma}' = \mathbf{J} \mathbf{W}(\mathbf{T}) \boldsymbol{\Sigma} \mathbf{W}^\top(\mathbf{T}) \mathbf{J}^\top$ and $\boldsymbol{\mu}' = \mathbf{J} \mathbf{W}(\mathbf{T}) \boldsymbol{\mu}$, respectively. Here, $\mathbf{W}(\mathbf{T})$ denotes the viewing transformation induced by $\mathbf{T}$, and $\mathbf{J}$ is the Jacobian of the affine approximation of the projective transformation.
For a given pixel $\mathbf{u}$, the final color $\hat{C}(\mathbf{u})$ and depth $\hat{D}(\mathbf{u})$ are computed by accumulating the contributions of overlapping, depth-sorted Gaussian primitives using standard $\alpha$-blending:
\begin{equation}
    \hat{{C}}(\mathbf{u}) = \sum_{i \in \mathcal{N}} \mathbf{c}_i \alpha_i \prod_{j=1}^{i-1} (1 - \alpha_j), \quad \hat{{D}}(\mathbf{u}) = \sum_{i \in \mathcal{N}} d_i \alpha_i \prod_{j=1}^{i-1} (1 - \alpha_j),
\end{equation}
where $\mathcal{N}$ denotes the ordered set of Gaussian primitives covering the pixel, and $d_i$ is the depth of the $i$-th Gaussian in the camera coordinate system. The effective 2D opacity is defined as $\alpha_i = \sigma_i \exp\!\left(-\frac{1}{2}(\mathbf{u} - \boldsymbol{\mu}_i')^\top {\boldsymbol{\Sigma}_i'}^{-1} (\mathbf{u} - \boldsymbol{\mu}_i')\right)$. By applying this pixel-wise accumulation, we obtain the rendered RGB image $\hat{\mathbf{I}}$ and depth map $\hat{\mathbf{D}}$. For notational convenience, the entire rendering pipeline is denoted as $\hat{\mathbf{I}}, \hat{\mathbf{D}} = \mathcal{R}(\mathcal{G}, \mathbf{T})$.
Optimization is performed by minimizing photometric loss $\mathcal{L}_P$ between the rendered image $\hat{\mathbf{I}}$ and the input image ${\mathbf{I}}$:
\begin{equation}
    \mathcal{L}_P = \lambda \|\mathbf{I} - \hat{\mathbf{I}}\|_1 + (1-\lambda)(1 - \text{SSIM}(\mathbf{I}, \hat{\mathbf{I}})),
\end{equation}
where $\|\cdot\|_1$ denotes the $L_1$ error, $\text{SSIM}$ is the Structural Similarity Index Measure~\cite{wang2004image}, and $\lambda$ is a hyperparameter. When depth observations are available, a depth loss is formulated as the $L_1$ error between the rendered and input depths.

\subsubsection{Deformable 3DGS}
Deformable 3DGS extends the static formulation to explicitly model temporal scene dynamics, making it particularly suitable for representing soft tissues~\cite{wu20244d,yang2024deformable,shan2025deformable,yang2024deform3dgs}. A prevalent approach adopts a {canonical-space-plus-deformation} paradigm. Here, the canonical space is parameterized by a set of static 3D Gaussians $\mathcal{G}^c = \left\{ \mathbf{G_i^c}\right\}_{i=1}^N$, while a deformation field $\mathcal{D}_t(\cdot \, ; \boldsymbol{\Theta}_t)$ maps these canonical Gaussians to their deformed states at time $t$:
\begin{equation}
    \mathbf{G}_i^t = \mathcal{D}_t(\mathbf{G}_i^c \, ; \boldsymbol{\Theta}_t),
\end{equation}
where $\mathbf{G}_i^t$ denotes the parameters of the $i$-th Gaussian in time $t$. The deformation field $\mathcal{D}_t$, parameterized by learnable parameters $\boldsymbol{\Theta}_t$, modulates Gaussian attributes, typically including position, orientation, and scale, over time. In practice, $\mathcal{D}_t$ can be instantiated using various function approximators, such as multilayer perceptrons (MLPs)~\cite{yang2024deformable}, multi-plane representations~\cite{wu20244d}, or Gaussian basis functions~\cite{shan2025deformable,yang2024deform3dgs}.

\subsection{System Overview}
\begin{figure}[t]
\centerline{\includegraphics[width=0.5\textwidth]{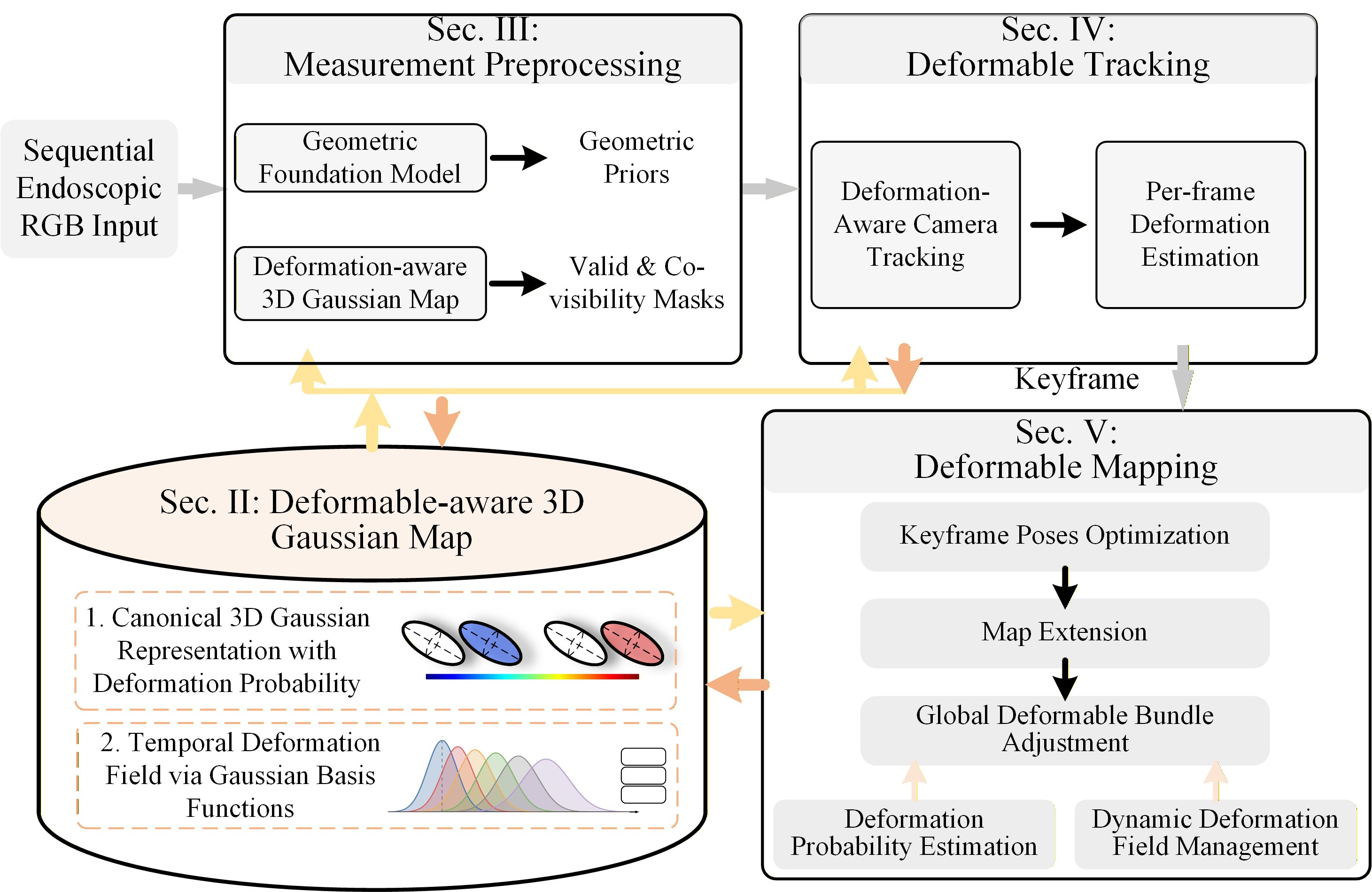}}
\caption{Overview of our proposed NRGS-SLAM. Central to our approach is the {deformation-aware 3D Gaussian map} (see Sec.~\ref{sec:map}), which represents the scene using canonical 3D Gaussians augmented by deformation probabilities and models temporal deformations via Gaussian Basis Functions. Given sequential RGB captured in a deformable environment, the {measurement preprocessing} module (see Sec.~\ref{sec:mea}) extracts geometric priors and generates valid and co-visibility masks. Subsequently, {deformable tracking} (see Sec.~\ref{sec:tracking}) performs frame-by-frame estimation of camera poses and scene deformations. Upon the insertion of a new keyframe, the {deformable mapping} module (see Sec.~\ref{sec:mapping}) is triggered to expand and globally optimize the map.}
    \label{fig:overview}
\end{figure}

Our system, NRGS-SLAM, jointly estimates camera motion and reconstructs photo-realistic representations of deformable endoscopic scenes from monocular video. The overall pipeline is illustrated in Fig.~\ref{fig:overview}. 
At the core of the system lies the {Deformation-aware 3D Gaussian Map} (Sec.~\ref{sec:map}). This map consists of a canonical 3D Gaussian representation augmented with per-primitive deformation probabilities, coupled with a temporal deformation field parameterized by Gaussian basis functions.
For each incoming frame, the pipeline begins with {Measurement Pre-processing} (Sec.~\ref{sec:mea}). This module uses geometric foundation models to derive essential geometric priors, including depth maps and sparse 2D/3D trajectories. It also generates masks for valid and co-visible pixels to ensure robust optimization.
Subsequently, the {Deformable Tracking} module (Sec.~\ref{sec:tracking}) performs frame-wise estimation of camera poses and scene deformations. It employs a coarse-to-fine tracking strategy that explicitly decouples camera ego-motion from scene deformation, followed by a per-frame optimization to capture the specific scene deformations observed in the current input frame.
Finally, {Deformable Mapping} (Sec.~\ref{sec:mapping}) operates at keyframe frequency to maintain and update the deformation-aware 3D Gaussian map. It expands the map by inserting new Gaussian primitives, together with their associated deformation probabilities and deformation fields, into newly observed regions. A global deformable bundle adjustment is then performed to jointly optimize keyframe poses and map parameters. This optimization is interleaved with dynamic deformation field management and a deformation probability estimation step, which generates supervisory signals for learning the deformation probabilities.

\section{Deformation-aware 3D Gaussian map}
\label{sec:map}

This section introduces deformation-aware 3D Gaussian map, which serves as the core representation of the proposed NRGS-SLAM system. 
In the following, we first detail the canonical space with deformation probability and then describe the temporal deformation field. The details regarding map expansion and optimization are presented in Sec.~\ref{sec:mapping}.

\subsection{Canonical Space with Deformation Probability}
\label{subsec:canonical_space}

\begin{figure}[t]
\centerline{\includegraphics[width=0.5\textwidth]{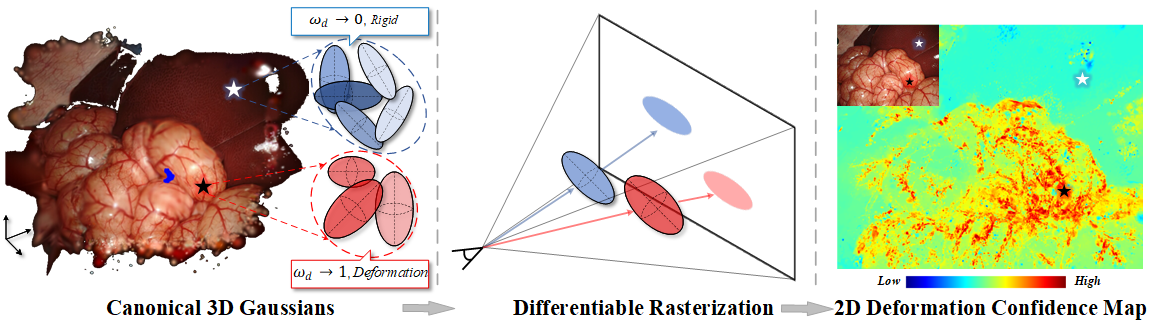}}
    \caption{ 
    \textbf{Left}: The canonical space is populated with 3D Gaussian primitives augmented by a learnable deformation probability $w_d$. Primitives in rigid regions (blue, $w_d \to 0$) remain static, while those in deformable tissues (red, $w_d \to 1$) are allowed to deform over time. 
    \textbf{Middle}: These probabilistic attributes are projected via differentiable rasterization.
    \textbf{Right}: The resulting 2D deformation confidence map $\mathbf{M}_{\text{def}}$ visualizes the pixel-wise confidence of scene non-rigidity. It serves as a crucial weighting mask to decouple camera ego-motion from intrinsic scene deformation during tracking and mapping.}
    \label{fig:def pro}
\end{figure}

Building upon the representation paradigm described in Sec.~\ref{sec:3dgs}, we represent the canonical space using a set of 3D Gaussian primitives $\mathcal{G}^c = \{\mathbf{G_i^c}\}_{i=1}^{N}$. 
To explicitly quantify the local rigidity of underlying structures, as shown in Fig.~\ref{fig:def pro}, we augment each canonical Gaussian with a learnable scalar attribute: the {deformation probability}, $w_{d,i} \in [0,1]$. 
This attribute is parameterized via a sigmoid function applied to an unconstrained logit $s_i \in \mathbb{R}$:
\begin{equation}
    w_{d,i} = \sigma(s_i).
    \label{eq:def_prob}
\end{equation}
The value $w_{d,i}$ acts as a continuous indicator of intrinsic deformability: $w_{d,i} \rightarrow 0$ denotes a rigid primitive, while $w_{d,i} \rightarrow 1$ indicates a fully deformable element. 
As this attribute is explicitly bound to each Gaussian primitive, we can leverage the differentiable rasterization pipeline to synthesize a dense 2D {deformation confidence map}, $\mathbf{M}_{\text{def}} \in [0,1]^{H \times W}$:
\begin{equation}
\mathbf{M}{_\text{def}}(\mathbf{u}) = \sum_{i \in \mathcal{N}(\mathbf{u})} w_{d,i}  \alpha_i \prod_{j=1}^{i-1} (1 - \alpha_j).
\end{equation}
By aggregating the intrinsic deformation probabilities along each viewing ray, $\mathbf{M}_{\text{def}}$ provides a pixel-wise confidence metric that distinguishes photometric variations caused by scene deformation from those induced by camera ego-motion.
Based on this distinction, the contributions of different image pixels can be adaptively modulated during optimization.
For instance, during camera pose estimation, pixels exhibiting high deformation likelihoods in $\mathbf{M}_{\text{def}}$ are down-weighted.
This strategy effectively anchors the tracking process to stable anatomical structures (see Sec.~\ref{sec:tracking}), thereby reducing the coupling ambiguity and enhancing tracking robustness.

Since ground-truth labels for tissue rigidity are unavailable in endoscopic scenarios, direct supervision of $w_{d,i}$ is impossible. Instead, we employ a Bayesian self-supervision strategy to derive a posterior deformation likelihood from monocular residuals. This posterior serves as a pseudo-ground-truth signal to guide the learning of deformation probabilities, as detailed in the optimization process in Sec.~\ref{sec:mapping}.

\subsection{Temporal Deformation Field via Gaussian Basis}
\label{subsec:deformation}

To capture the temporal evolution of scene, we extend the canonical 3D Gaussians into the time domain using a learnable deformation field. Furthermore, we incorporate deformation probability to modulate this field, ensuring that deformations are applied selectively based on structural rigidity.

\subsubsection{1D Gaussian Basis Functions}
In contrast to implicit deformation networks (e.g., MLPs), we parameterize the temporal variation of each Gaussian using 1D Gaussian basis functions. This explicit representation facilitates dynamic map management (see Sec.~\ref{sec:mapping}). Specifically, each Gaussian $i$ is assigned a set of basis functions to modulate its attributes at time $t$.
Let $\phi$ denote a single basis function, parameterized by a temporal center $\tau$ and a temporal extent $\sigma$:
\begin{equation}
    \phi(t; \tau, \sigma) = \exp\left(-\frac{(t - \tau)^2}{2\sigma^2}\right).
\end{equation}
The temporal offset for an attribute $\mathcal{A} \in \{\boldsymbol{\mu}, \mathbf{s}, \mathbf{q}\}$ of the $i$-th Gaussian is defined as a weighted sum of $K^{\mathcal{A}}$ basis functions:
\begin{equation}
    \Delta \mathcal{A}_i(t) = \sum_{k=1}^{K^{\mathcal{A}}} \omega_{i,k}^{\mathcal{A}} \cdot \phi(t; \tau_{i,k}^{\mathcal{A}}, \sigma_{i,k}^{\mathcal{A}}),
\end{equation}
Here, $\omega_{i,k}^{\mathcal{A}}$, $\tau_{i,k}^{\mathcal{A}}$, and $\sigma_{i,k}^{\mathcal{A}}$ denote the learnable weight, temporal center, and temporal extent of the $k$-th basis function, respectively. We collectively denote these parameters as $\boldsymbol{\Theta}_t^{\mathcal{A}}$.

\subsubsection{Deformation-probability modulation} 
To prevent spurious motion in static regions, we employ the deformation probability $w_{d,i}$ as a {soft gating} mechanism. Formally, the attributes of the $i$-th Gaussian at time $t$ are derived by adding the probability-weighted offsets to the canonical state:
\begin{equation}
    \mathcal{A}_i(t) = \mathcal{A}_i^c + w_{d,i} \cdot \Delta \mathcal{A}_i(t), \quad \forall \mathcal{A} \in \{\boldsymbol{\mu}, \mathbf{s}, \mathbf{q}\},
    \label{equ:pro_def}
\end{equation}
where $\mathcal{A}_i^c$ denotes the canonical attribute of the $i$-th Gaussian. For the rotation quaternion $\mathbf{q}$, an additional normalization step is performed post-update to enforce the unit norm constraint.
This formulation promotes physically plausible deformation: transformations are suppressed in rigid regions (where $w_{d,i} \to 0$), effectively anchoring them to the canonical geometry, whereas temporal basis functions remain fully active in deformable tissues (where $w_{d,i} \to 1$).
For brevity, we denote the transformation of the canonical Gaussian set $\mathcal{G}^c$ to its state at time $t$ as $\mathcal{G}^t = \mathcal{D}_{\boldsymbol{\Theta}_t}(\mathcal{G}^c, t)$.
\section{Measurement Preprocessing}
\label{sec:mea}

\begin{figure}[t]
    \centering
    \includegraphics[width=\linewidth]{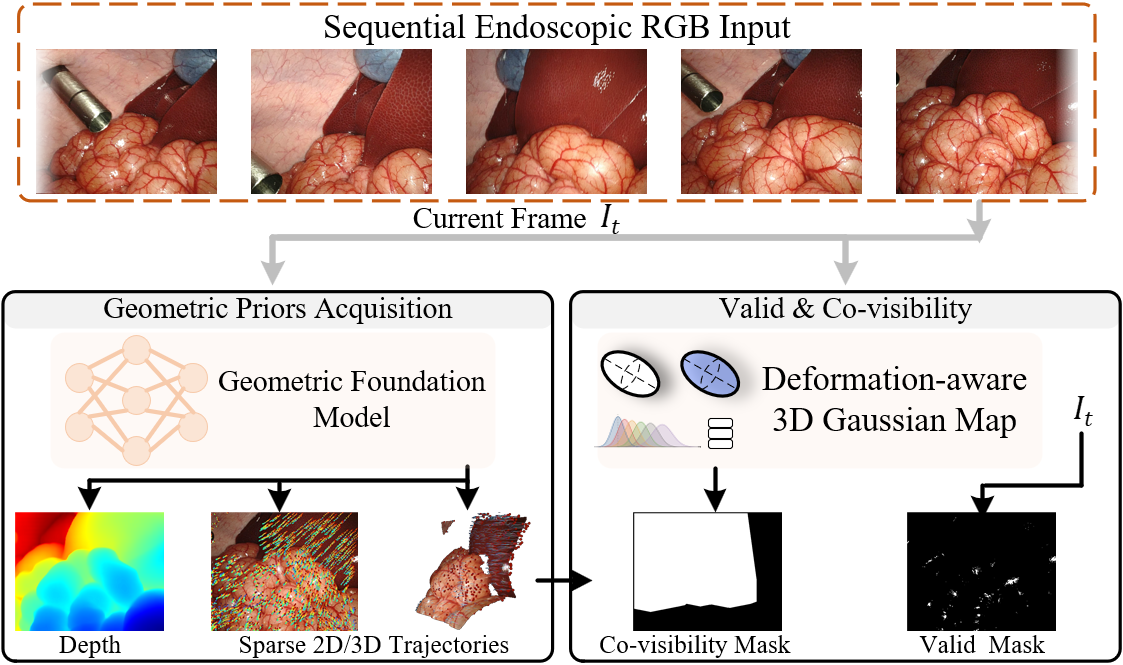}
    \caption{Overview of the measurement preprocessing pipeline. The module processes the incoming frame $I_t$ to extract geometric cues via a geometric foundation model (left) and determines valid and co-visibility masks (right).}
    \label{fig:mea}
\end{figure}

This section details the preprocessing of monocular visual data for subsequent deformable tracking and mapping. As illustrated in Fig.~\ref{fig:mea}, the process focuses on extracting geometric priors and generating validity and co-visibility masks for the current frame.
Formally, at time step $t$, we take an RGB frame $\mathbf{I}_t$ as input. We assume that the deformation-aware 3D Gaussian map from the previous step, comprising a canonical Gaussian set $\mathcal{G}^c$ and a temporal deformation field $\mathcal{D}_{\boldsymbol{\Theta}_{t-1}}$, is available.

\subsection{Geometric Priors Acquisition}
\label{subsec:geo_prior}

Leveraging recent advances in geometric foundation models, we incorporate learned geometric priors to alleviate the ill-posed nature of monocular non-rigid SLAM.
Specifically, upon receiving frame $\mathbf{I}_t$, a geometric foundation model jointly predicts a scale-consistent depth map $\mathbf{D}_t$, sparse 2D point trajectories $\mathbf{T}^{2d}_t$, and corresponding 3D trajectories $\mathbf{T}^{3d}_t$.
These trajectories are defined as $\mathbf{T}^{2d}_t = \{ (\mathbf{u}^{t_0}_i, \mathbf{u}^t_i) \}_{i \in \mathcal{F}_t}$ and $\mathbf{T}^{3d}_t = \{ (\mathbf{x}^{t_0}_i, \mathbf{x}^t_i) \}_{i \in \mathcal{F}_t}$, where $t_0$ represents the starting timestamp and $\mathcal{F}_t$ denotes the set of tracked pixel indices at time $t$.
The 3D and 2D trajectories are aligned, sharing identical indices and temporal spans.
The extracted priors at time $t$ are summarized as:
$
\mathcal{P}_t =
\left\{
\mathbf{D}_t,\ 
\mathbf{T}^{2d}_{t_0 \rightarrow t},\ 
\mathbf{T}^{3d}_{t_0 \rightarrow t}
\right\}.
$

We employ SpatialTrackerV2~\cite{xiao2025spatialtrackerv2} for this task, though our framework is model-agnostic and compatible with other methods providing similar cues.
While these priors offer vital geometric constraints, they contain inherent noise due to model inaccuracies, scene dynamics, and the domain gap in endoscopic imagery.
To handle this, we design a robust strategy to exploit these priors while mitigating noise, as detailed in Sec.~\ref{sec:tracking}.

\subsection{Validity and Co-visibility Masks Determination}
\label{subsec:covisibility}

Endoscopic procedures typically rely on a camera-mounted light source, which frequently induces significant illumination variations across the image. This results in under-exposed or over-saturated regions where the photometric consistency assumption is violated. To mitigate this, we employ an intensity-based validity mask $\mathbf{M}_t^{\text{val}}$, to exclude such outliers. For a pixel $\mathbf{u}$ in the current frame $\mathbf{I}_t$, the mask is defined as:
\begin{equation}
\mathbf{M}_t^{\text{val}}(\mathbf{u}) =
\begin{cases}
1, & \delta \le G_t(\mathbf{u}) \le 1 - \delta, \\
0, & \text{otherwise}.
\end{cases}
\end{equation}
where $G_t(\mathbf{u}) \in [0, 1]$ denotes the normalized gray-scale intensity and $\delta$ is a predefined threshold.

Subsequently, we identify the co-visible region where the current observation aligns with the existing scene representation. This is critical, as unrepresented areas appear transparent in rendered views and can introduce optimization artifacts. We determine the co-visible region by intersecting a map-based mask ($\mathbf{M}_t^{\text{map}}$) with a 2D trajectory mask ($\mathbf{M}_t^{\text{track}}$). Specifically, we first compute $\mathbf{M}_t^{\text{map}}$ to identify pixels covered by the Gaussian map at time $t$. By approximating the current state via temporal continuity ($\boldsymbol{\Theta}_t \approx \boldsymbol{\Theta}_{t-1}$), we render a transmittance map $\hat{\mathcal{T}}_t$ from the predicted Gaussian map $\mathcal{G}^t = \mathcal{D}_{\boldsymbol{\Theta}_t}(\mathcal{G}^c, t)$. Since regions lacking Gaussian primitives result in high transmittance, we effectively isolate the occupied space as:
\begin{equation}
\hat{\mathcal{T}}_t(\mathbf{u}) = \prod_{i \in \mathcal{N}(\mathbf{u})} \left(1 - \alpha_i\right), \quad
\mathbf{M}_t^{\text{map}}(\mathbf{u}) = \mathbb{I}\!\left[\hat{\mathcal{T}}_t(\mathbf{u}) < \tau_{\mathcal{T}}\right],
\end{equation}
where $\tau_{\mathcal{T}}$ is the transparency threshold. Simultaneously, we derive $\mathbf{M}_t^{\text{track}}$ by tracking sparse 2D features across frames and applying an $\alpha$-shape algorithm~\cite{opencv_library} to generate a boundary that physically constrains the trackable region. The final co-visibility mask is obtained by the intersection:
\begin{equation}
\mathbf{M}_t^{\text{covis}}(\mathbf{u}) =
\mathbf{M}_t^{\text{track}}(\mathbf{u}) \cdot
\mathbf{M}_t^{\text{map}}(\mathbf{u}).
\end{equation}
\section{Deformable Tracking}
\label{sec:tracking}

\begin{figure}[t]
\centerline{\includegraphics[width=0.5\textwidth]{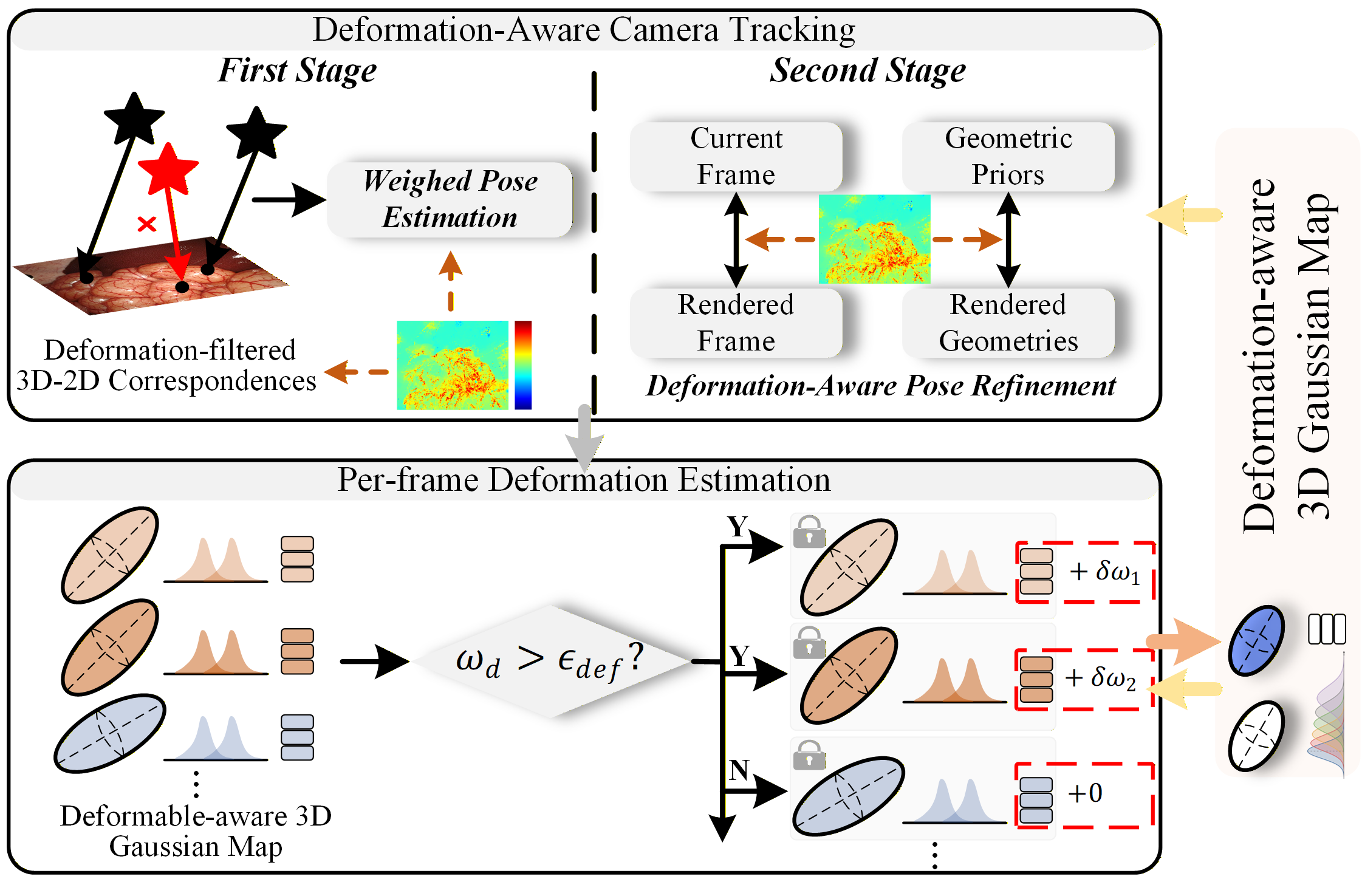}}
\caption{Overview of the deformable tracking pipeline. 
\textbf{Top: Deformation-Aware Camera Tracking} estimates the camera pose in a coarse-to-fine manner. 
The first stage computes an initial pose using deformation-filtered 3D-2D correspondences. 
The second stage refines this pose by aligning rendered frame and geometries with current observations and geometric priors; crucially, this optimization is {weighted by the rendered deformation probability map} to prioritize reliable rigid regions.
\textbf{Bottom: Per-frame Deformation Estimation} updates the deformation field to match the input frame. 
Guided by the deformation probabilities, we employ an efficient residual-based optimization to selectively capture non-rigid variations.}
\label{fig:tracking}
\end{figure}

This section details the deformable tracking module, which aims to estimate the current camera pose $\mathbf{T}_t$ and the associated deformation parameters $\boldsymbol{\theta}_t$ based on the deformation-aware 3D Gaussian representation. As shown in Fig.~\ref{fig:tracking}, the tracking pipeline consists of a two-stage deformation-aware camera pose estimation followed by a per-frame deformation update. We first formulate the objective functions governing the tracking process and then describe each stage in detail.

\subsection{Objective Functions}
\label{subsec:obj fun}
We formulate the tracking objective by combining a photometric loss with a robust geometric loss, leveraging the raw RGB input $\mathbf{I}_t$ and a set of geometric priors $\mathcal{P}_t$.

\subsubsection{Photometric Loss}
The photometric loss is defined as the $\ell_2$ error between the rendered RGB and the input RGB:
\begin{equation}
\mathcal{L}_{\text{ph}}
=
\frac{1}{|\Omega|}
\sum_{\mathbf{u} \in \Omega}
\mathbf{M}_t^{\text{val}}(\mathbf{u}) \,
\mathbf{M}_t^{\text{covis}}(\mathbf{u}) \,
\left\|
\hat{\mathbf{I}}_t(\mathbf{u}) -
\mathbf{I}_t(\mathbf{u})
\right\|_2^2 .
\label{eq:photo_loss}
\end{equation}
Here, $\hat{\mathbf{I}}_t(\mathbf{u})$ and $\mathbf{I}_t(\mathbf{u})$ denote the rendered and observed RGB values at pixel $\mathbf{u}$, respectively.
The validity and co-visibility masks, $\mathbf{M}_t^{\text{val}}$ and $\mathbf{M}_t^{\text{covis}}$, derived in Sec.~\ref{sec:mea}, restrict the loss to reliable pixels and regions explainable by the current map.

Note that although this photometric loss underlies both camera pose tracking and per-frame deformation estimation, its formulation differs across stages, especially regarding the integration of deformation probability maps. The detailed stage-specific definitions are provided in the next subsection.

\subsubsection{Geometric Loss}
We adopt a robust geometric loss that unifies multiple types of geometric constraints while reducing the influence of unreliable priors. We first present a unified formulation of the geometric loss and then describe how the quantities associated with each geometric prior are rendered from the current estimates of the camera pose and scene map. Specifically, at the $k$-th optimization iteration, the geometric supervision is formulated as a robust optimization problem, in which the geometric loss is defined as a weighted sum of individual geometric constraint terms:
\begin{equation}
\mathcal{L}_{\mathrm{geo}}^{(k)}
=
\sum_{g \in \mathcal{P}_t}
\lambda_g(k)\,
\mathcal{L}_g^{(k)},
\label{eq:geo_total}
\end{equation}
where $g$ indexes the specific geometric prior in $\mathcal{P}_t = \{\mathbf{D}_t, \mathbf{T}^{2d}_{t_{0}\rightarrow t}, \mathbf{T}^{3d}_{t_{0}\rightarrow t}\}$.
The weight $\lambda_g(k)$ follows an exponential annealing schedule: $\lambda_g(k) = \lambda_g^{0} \exp(-k / \tau_g) + \lambda_g^{\min}$, where $\lambda_g^{0}$ and $\lambda_g^{\min}$ are the initial and minimum weights, and $\tau_g$ controls the decay rate. This schedule prioritizes geometric supervision in the early phases to ensure convergence, progressively reducing its influence as the solution stabilizes.
Formally, the loss term $\mathcal{L}_g^{(k)}$ for a specific prior $g$ is defined via Iteratively Reweighted Least Squares (IRLS)~\cite{holland1977robust}:
\begin{equation}
\begin{aligned}
\mathcal{L}_g^{(k)}
&=
\frac{1}{|\mathcal{U}_g|}
\sum_{\mathbf{u} \in \mathcal{U}_g}
\mathbf{M}_t^{\text{val}}(\mathbf{u}) \,
\mathbf{M}_t^{\text{covis}}(\mathbf{u}) \,
\frac{1}{2}
\gamma^{(k)}_{g}(\mathbf{u})
\bigl(r^{g}_t(\mathbf{u})\bigr)^2, \\
\gamma^{(k)}_{g}(\mathbf{u})
&=
\frac{
\rho'\!\left(r^{g,(k)}_t(\mathbf{u})\right)
}{
r^{g,(k)}_t(\mathbf{u}) + \epsilon
},
\,
r^{g}_t(\mathbf{u})
=
\left\|
\hat{\mathbf{G}}^{g}_t(\mathbf{u})
-
\mathbf{G}^{g}_t(\mathbf{u})
\right\|_2.
\end{aligned}
\label{eq:geo_each}
\end{equation}
Here, the domain $\mathcal{U}_g$ varies by prior type: $\mathcal{U}_{\mathbf{D}} = \Omega$ corresponds to the image domain for the dense depth map, while $\mathcal{U}_{\mathbf{T}^{2d}} = \mathcal{U}_{\mathbf{T}^{3d}} = \mathcal{F}_t$ denotes the set of tracked sparse feature points.
The weight $\gamma^{(k)}_{g}(\mathbf{u})$ is derived from a robust penalty function $\rho(\cdot)$ (with derivative $\rho'(\cdot)$), where $\epsilon$ is a small constant for numerical stability.
The residual $r^{g}_t(\mathbf{u})$ measures the $\ell_2$ error between the observed prior $\mathbf{G}^{g}_t(\mathbf{u})$ and its predicted counterpart $\hat{\mathbf{G}}^{g}_t(\mathbf{u})$. In the following, we describe how these predicted quantities are rendered from the current scene representation and camera pose.

The depth map is rendered using the same differentiable Gaussian splatting pipeline as the RGB image. Given the canonical Gaussian map $\mathcal{G}^c$ and the deformation parameters $\boldsymbol{\theta}_t$, the depth map at time $t$ is rendered under the current camera pose $\mathbf{T}_t$ as
$\hat{\mathbf{D}}_t = \mathcal{R}\!\left(\mathcal{D}_{\boldsymbol{\theta}_t}(\mathcal{G}^c, t), \mathbf{T}_t\right)$.
In addition, our deformation-aware 3D Gaussian map enforces temporally consistent motion for individual Gaussians within the canonical space. Leveraging this property, we establish pixel-wise temporal correspondences by aggregating Gaussian motions, following a strategy similar to~\cite{wang2023tracking}. Specifically, for each pixel $\mathbf{u} \in \Omega$ in frame $\mathbf{I}_{t_0}$, we use $\mathcal{H}(\mathbf{u})$ to denote the ordered set of Gaussians in the canonical space contributing to pixel $\mathbf{u}$ in the splatting process.
To track the surface point corresponding to $\mathbf{u}$ from time $t_0$ to $t$, we compute the expected displacement by accumulating the motion of each Gaussian in $\mathcal{H}(\mathbf{u})$, weighted by their opacity-based contributions:
\begin{equation}
\hat{\mathbf{G}}^{T^{3d}}_{t_0 \rightarrow t}(\mathbf{u})
=
\sum_{i \in \mathcal{H}(\mathbf{u})}
\left(
\boldsymbol{\mu}_i^{t}
-
\boldsymbol{\mu}_i^{t_0}
\right)
\alpha_i
\prod_{j=1}^{i-1}
\left( 1 - \alpha_j \right).
\end{equation}
Here, $\boldsymbol{\mu}_i^{t_0}$ and $\boldsymbol{\mu}_i^{t}$ denote the centers of the $i$-th Gaussian at times $t_0$ and $t$, respectively,obtained by applying the deformation field defined in Eq.~\eqref{equ:pro_def} to the canonical Gaussian center.

The corresponding 2D trajectory is obtained by projecting the predicted 3D position into the camera frame at time $t$:
\begin{equation}
\hat{\mathbf{G}}^{T^{2d}}_{t_1 \rightarrow t_2}(\mathbf{u})
=
\Pi\!\left(
\mathbf{K}
\mathbf{T}_{t}
\begin{bmatrix}
\hat{\hat{\mathbf{G}}}^{T^{3d}}_{t_0 \rightarrow t}(\mathbf{u})
1
\end{bmatrix}
\right),
\end{equation}
where $\Pi(\cdot)$ is the perspective projection, $\mathbf{K}$ is the intrinsic matrix, and $\mathbf{T}_{t}$ is the camera pose.

\subsection{Deformation-Aware Camera Tracking}
In deformable endoscopic environments, pixel variations arise from the combined effects of camera ego-motion and intrinsic tissue deformation, which degrade tracking accuracy and stability. To address this challenge, we leverage the deformation probability map derived from the deformation-aware 3D Gaussian map (Sec.~\ref{sec:map}) and carefully design a robust coarse-to-fine camera tracking strategy. In the first stage, we estimate an initial camera pose by solving a deformation-weighted Perspective-$n$-Point (PnP) problem using filtered sparse correspondences. In the second stage, we refine the pose by jointly enforcing photometric and geometric consistency. Details of these two stages are described below.

\subsubsection{Deformation-Weighted Coarse Pose Estimation}
In this stage, given the canonical Gaussian map $\mathcal{G}^c$ and the deformation parameters $\boldsymbol{\theta}_{t_{ak}}$ associated with the adjacent keyframe $\mathbf{I}_{ak}$, we render the depth map $\hat{\mathbf{D}}_{ak}$ together with the deformation confidence map $\mathcal{M}_{\mathrm{def}}$:
$\hat{\mathbf{D}}_{ak}, \mathcal{M}_{\mathrm{def}}^{ak} =
\mathcal{R}\!\left(
\mathcal{D}_{\boldsymbol{\theta}_{t_{ak}}}(\mathcal{G}^c, t_{ak}),
\mathbf{T}_{t_{ak}}
\right)$,
where all involved parameters are pre-optimized.
In parallel, we sample pixels from the keyframe $\mathbf{I}_{ak}$ and use  the geometric foundation model~\cite{xiao2025spatialtrackerv2} to predict their 2D trajectories $\mathbf{T}^{2d}_{t_{ak}\rightarrow t} = \{(\mathbf{u}^{ak}_i, \mathbf{u}^{t}_i)\}_{i=1}^{N}$, establishing correspondences between the keyframe $\mathbf{I}_{ak}$ and the current frame $\mathbf{I}_t$.
These 2D trajectories are used only to initialize the current camera pose and do not affect subsequent stages.
To ensure robustness, we retain only trajectories satisfying $\mathbf{M}_{\text{def}}^{ak}(\mathbf{u}^{ak}_i) < \tau_{\mathrm{def}}$, thereby filtering out correspondences in regions with high deformation probability.
For each retained trajectory, the pixel $\mathbf{u}^{ak}_i$ is back-projected into 3D using the rendered depth
$z_i = \hat{\mathbf{D}}_{ak}(\mathbf{u}^{ak}_i)$ and the known keyframe pose $\mathbf{T}_{t_{ak}}$, yielding a 3D point
$\mathbf{X}^w_i$ in the world coordinate system.
This process generates a set of 3D--2D correspondences $\{(\mathbf{X}^w_i, \mathbf{u}^t_i)\}$.
We then estimate the initial camera pose $\mathbf{T}_{t}$ by solving a weighted PnP problem:
\begin{equation}
    \min_{\mathbf{T}_{t}}
    \sum_i w_i
    \left\|
    \mathbf{K}\,\mathbf{T}_{t}\,\mathbf{X}^w_i
    -
    \mathbf{u}^{t}_i
    \right\|_2 .
\end{equation}
Each correspondence is weighted by its deformation confidence,
$w_i = \left(1 - \mathcal{M}_{\mathrm{def}}^{ak}(\mathbf{u}^{ak}_i)\right)^2$,
which reduces the influence of residual non-rigid motion and biases the estimation toward stable structures. The optimization is solved using a PnP solver within a RANSAC scheme~\cite{fischler1981random, lepetit2009ep}, yielding a robust initial pose estimate $\mathbf{T}_t$.

\subsubsection{Deformation-Aware Pose Refinement}

In the second stage, we refine the camera pose $\mathbf{T}_t$ estimated in the previous stage by leveraging dense photometric observations and geometric priors. Recall that given the deformable scene representation, including the canonical Gaussian map $\mathcal{G}^c$ and the deformation parameters $\boldsymbol{\theta}_t$, together with the coarse camera pose $\mathbf{T}_t$, we render RGB, depth, and 3D--2D trajectories (see Sec.~\ref{subsec:obj fun}).  
At this stage, the deformation field is approximated using the parameters from the previous timestamp, i.e., $\boldsymbol{\theta}_t \approx \boldsymbol{\theta}_{t-1}$. We further incorporate a deformation-aware weighting scheme based on the rendered deformation confidence map $\mathcal{M}_{\mathrm{def}}^t$, which is rendered in the current frame $\mathbf{I}_t$ using the coarse pose estimate $\mathbf{T}_t$.
The objective function for pose refinement is defined as:
\begin{equation}
    \min_{\mathbf{T}_{t}} \quad
    \sum_{\mathbf{u}}
    \underbrace{\left( 1 - \mathcal{M}_{\mathrm{def}}^t(\mathbf{u}) \right)}_{\text{Rigidity Weight}}
    \cdot
    \left( 
        \mathcal{L}_{\text{ph}}(\mathbf{u}, \mathbf{T}_t) 
        + 
        \lambda_{\text{geo}} \mathcal{L}_{\text{geo}}(\mathbf{u}, \mathbf{T}_t)
    \right).
    \label{eq:refine_total}
\end{equation}
Here, $\mathcal{L}_{\text{ph}}$ and $\mathcal{L}_{\text{geo}}$ denote the photometric and geometric losses, respectively. The term $(1 - \mathcal{M}_{\mathrm{def}}^t(\mathbf{u}))$ acts as a rigidity confidence weight, reducing the influence of regions with high deformation probability. As a result, the refinement process is guided mainly by rigid or low-deformation regions, which provide more reliable constraints for accurate pose estimation.

\subsection{Per-frame Deformation Estimation}
\label{subsec:def_estimation}

Subsequent to camera pose estimation, the deformation field is updated to align the scene representation with the non-rigid deformations observed in the current image frame.
While optimizing all deformation parameters is theoretically possible, it is computationally intractable for real-time applications.
Therefore, we employ an efficient residual-based optimization strategy.
Specifically, we retain the temporal basis parameters $\tau_{i,k}^{\mathcal{A}}$ and $\sigma_{i,k}^{\mathcal{A}}$ as fixed priors, which define the global temporal structure learned from the adjacent keyframe.
We then solve for the deformation at the current time $t$ by optimizing only a time-dependent residual $\delta \omega_{i,k}^{\mathcal{A},(t)}$ added to the basis weights.
The resulting temporal offset for an attribute $\mathcal{A} \in \{\boldsymbol{\mu}, \mathbf{s}, \mathbf{q}\}$ of the $i$-th Gaussian is formulated as:
\begin{equation}
    \Delta \mathcal{A}_i(t)
    =
    \sum_{k=1}^{K^{\mathcal{A}}}
    \left(
        \omega_{i,k}^{\mathcal{A}}
        +
        \delta \omega_{i,k}^{\mathcal{A},(t)}
    \right)
    \cdot
    \phi\!\left(
        t;\,
        \tau_{i,k}^{\mathcal{A}},
        \sigma_{i,k}^{\mathcal{A}}
    \right).
    \label{eq:residual_offset}
\end{equation}
This formulation enables the model to capture frame-specific surface variations while preserving the structural consistency established at the keyframe.

To further reduce computational cost, we leverage the deformation probability map introduced in Sec.~\ref{subsec:canonical_space}.
We assume that Gaussian primitives classified as rigid in the canonical space do not require deformation updates.
As a result, residuals are optimized only for Gaussians whose deformation probability $w_{d,i}$ exceeds a threshold $\epsilon_{\text{def}}$:
\begin{equation}
    \delta \omega_{i,k}^{\mathcal{A},(t)} =
    \begin{cases}
        \text{optimized}, & \text{if } w_{d,i} > \epsilon_{\text{def}}, \\
        0, & \text{otherwise}.
    \end{cases}
\end{equation}
This masking strategy suppresses unnecessary updates on rigid structures and focuses the optimization on deformable regions.

Finally, we formulate the deformation update as a minimization problem.
Let $\delta \boldsymbol{\omega}_t$ denote the vector of all active residual coefficients at frame $t$.
The objective function combines the photometric and geometric losses (Sec.~\ref{subsec:obj fun}) with two regularization terms:
\begin{equation}
    \min_{\delta \boldsymbol{\omega}_t} \quad
    \mathcal{L}_{\text{ph}}
    +
    \mathcal{L}_{\text{geo}}
    +
    \lambda_{\text{reg}}
    \|\delta \boldsymbol{\omega}_t\|_2^2
    +
    \lambda_{\text{tem}}
    \|\delta \boldsymbol{\omega}_t -
      \delta \boldsymbol{\omega}_{t-1}\|_2^2.
    \label{eq:def_optimization}
\end{equation}
The term weighted by $\lambda_{\text{reg}}$ limits the magnitude of the residuals, preventing excessive deviation from the base weights and ensuring physical plausibility.
The final term enforces temporal smoothness by penalizing abrupt changes in deformation between consecutive frames.

\subsection{New Keyframe Selection}
\label{subsec:key_sel}

The following criteria are used to determine whether the current frame $\mathbf{I}_t$ should be selected as a new keyframe.

\textbf{Insufficient Map Co-visibility.}
We measure the overlap between the current observation and the scene map using the co-visibility mask $\mathbf{M}_t^{\text{covis}}$ (Sec.~\ref{subsec:covisibility}).
A co-visible pixel ratio below $0.75$ indicates that the current view contains newly explored areas or is largely occluded.
Consequently, a new keyframe is inserted.

\textbf{Large Deformation Magnitude.}
The residual-based estimation described in Sec.~\ref{subsec:def_estimation} assumes that non-rigid motion in the current frame can be modeled as a moderate perturbation of the basis weights of the reference keyframe.
When the optimized residuals $\delta \omega_{i,k}^{\mathcal{A},(t)}$ become large relative to the original coefficients, this assumption becomes less valid. We quantify this deviation using the relative residual ratio:
\begin{equation}
    R_{\text{def}}(t)
    =
    \frac{1}{|\mathcal{G}_d|}
    \sum_{i \in \mathcal{G}_d}
    \sum_{k}
    \frac{
        \left\|
        \delta \omega_{i,k}^{\boldsymbol{\mu},(t)}
        \right\|_2
    }{
        \left\|
        \omega_{i,k}^{\boldsymbol{\mu}}
        \right\|_2 + \epsilon
    },
\end{equation}
where $\mathcal{G}_d = \{ i \mid w_{d,i} > \epsilon_{\text{def}} \}$ and $\epsilon$ is a constant for numerical stability.
We compute $R_{\text{def}}(t)$ using only position-related residuals ($\mathcal{A}=\boldsymbol{\mu}$), as they more directly reflect non-rigid structural changes. A new keyframe is selected when $R_{\text{def}}(t)$ exceeds a predefined threshold 0.1.

\textbf{Significant Camera Motion.}
We monitor the relative translation between the current camera pose and the most recent keyframe.
A new keyframe is selected if this translation exceeds a predefined distance threshold $8mm$.

\textbf{Maximum Frame Interval.}
To limit drift accumulation over long sequences, we impose a temporal constraint.
A new keyframe is automatically inserted if the number of frames since the last keyframe exceeds 20.
\section{Deformable Mapping}
\label{sec:mapping}

\begin{figure*}[th]
\centerline{\includegraphics[width=1\textwidth]{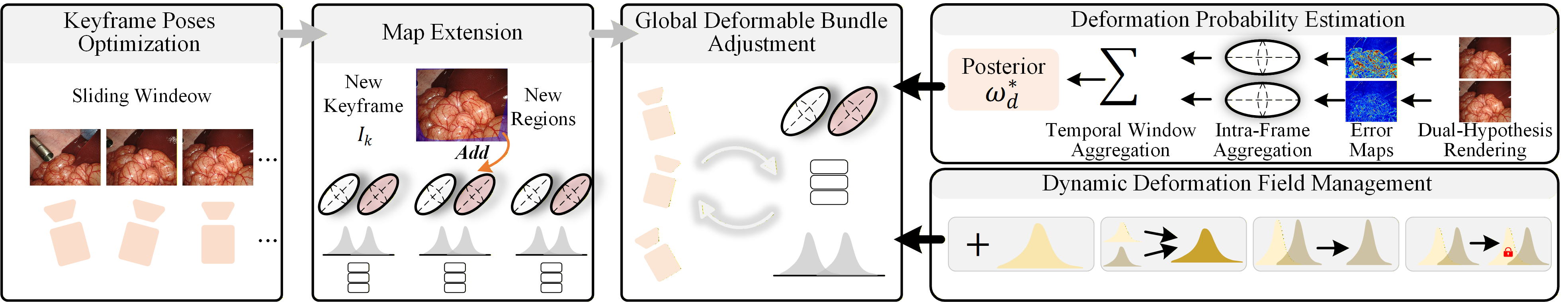}}
\caption{
Overview of the deformable mapping.
When a new keyframe is inserted, the system operates within a sliding window to optimize camera poses and extend the map with newly observed regions.
A global deformable bundle adjustment jointly refines camera states and scene parameters.
Within this optimization, the Deformation Probability Estimation module (Sec.~\ref{subsec:prob_estimation}) infers posterior deformation probabilities to supervise their optimization, while the Dynamic Deformation Field Management module (Sec.~\ref{subsec:dynamic_management}) adaptively adjusts the number of temporal basis functions.
}
\label{fig:mapping}
\end{figure*}

This section details the deformable mapping module, which is responsible for maintaining and refining the global scene representation. Following previous works, we employ a sliding window of keyframes, denoted as $\mathcal{W}_t$, to ensure computational efficiency and local consistency.
When the current frame $t$ is selected as a new keyframe, it is appended to $\mathcal{W}_t$, and the oldest keyframe is marginalized to maintain a fixed window size of $M$.
We define the canonical Gaussian map as $\mathcal{G}^c = \{\mathbf{G_i^c}, w_{d,i}\}$ and the deformation field as $\mathcal{D}_{\boldsymbol{\theta}_t}$, parameterized by $\boldsymbol{\theta}_t$.
As illustrated in Fig.~\ref{fig:mapping},
the pipeline executes three sequential steps: 
1) optimizing the poses of keyframes within $\mathcal{W}_t$; 
2) extending $\mathcal{G}^c$ to encompass newly observed regions; 
and 3) performing a global deformable bundle adjustment. 
Crucially, the bundle adjustment phase incorporates a Bayesian self-supervision strategy to guide the optimization of deformation probabilities and dynamically manage deformation field.

\subsection{Keyframe Poses Optimization}
\label{subsec:keyframe_pose_opt}

To mitigate the accumulation of tracking drift, we perform pose optimization over the sliding window of keyframes $\mathcal{W}_t$ through a small number of iterations. 
This process extends the frame-to-map tracking formulation (Sec.~\ref{sec:tracking}) to a multi-frame setting. 
Specifically, for each keyframe $k \in \mathcal{W}_t$, we render the deformation confidence map $\mathbf{M}_{\text{def}}^k$, along with the RGB, depth, and 2D/3D trajectories from the current map. 
The objective function minimizes the weighted photometric and geometric losses aggregated over the window:
\begin{equation}
    \min_{ \{ \mathbf{T}_{k} \} } 
    \sum_{k \in \mathcal{W}_t}
    \sum_{\mathbf{u}}
    \left( 1 - \mathcal{M}_{\mathrm{def}}^k(\mathbf{u}) \right)
    \left( 
        \mathcal{L}_{\text{ph}}(\mathbf{u}, \mathbf{T}_k) 
        + 
        \lambda_{\text{geo}} \mathcal{L}_{\text{geo}}(\mathbf{u}. \mathbf{T}_k)
    \right),
    \label{eq:sliding_window_pose}
\end{equation}
where $\{ \mathbf{T}_{k} \}$ denotes the set of camera poses to be optimized. 
During this stage, we hold all map-related parameters fixed.

\subsection{Map Extension}
\label{subsec:map_extension}

We dynamically expand the map to accommodate newly observed regions whenever a new keyframe is inserted.
We identify candidate regions for expansion by utilizing $\hat{\mathcal{T}}_t(\mathbf{u})$ and $\mathbf{M}_t^{\text{track}}(\mathbf{u})$ defined in Sec.~\ref{subsec:covisibility}. 
Unlike the co-visibility check, which validates regions already covered by the existing representation, map extension targets the complementary areas. 
Specifically, a pixel $\mathbf{u}$ is considered a valid target for generation if it exhibits high transmittance, indicating free space along the ray, and lies outside the support of tracked sparse features:
$
    \mathbf{M}_{\text{new}}(\mathbf{u}) 
    = \mathbb{I}\!\left[ \hat{\mathcal{T}}_t(\mathbf{u}) \ge 0.1 \right]
    \cdot \left( 1 - \mathbf{M}_t^{\text{track}}(\mathbf{u}) \right).
$

For the identified regions, we add new elements to the deformation-aware 3D Gaussian map.  
In the canonical space, Gaussian attributes, including the mean, rotation, scale, opacity, and spherical harmonics coefficients, are initialized following standard practices~\cite{kerbl20233d,matsuki2024gaussian}. The deformation probability is initialized with a prior value of $w_{d,i} = 0.6$.
In addition, the temporal basis functions of these Gaussians are aligned with their insertion time $t$. Specifically, the temporal centers are set to $\tau_{i,k}^{\mathcal{A}} = t$, the extents are initialized to the mean temporal extent of the existing basis functions and denoted as $\sigma_{i,k}^{\mathcal{A}}$, and the basis weights are set to $\omega_{i,k}^{\mathcal{A}} = 0$.

\subsection{Global Deformable Bundle Adjustment}
\label{subsec:global_ba}

Inspired by state-of-the-art SLAM systems~\cite{rodriguez2024nr}, we perform global deformable bundle adjustment to jointly optimize the canonical Gaussian primitives $\mathcal{G}^c$, the temporal deformation field $\mathcal{D}_{\boldsymbol{\theta}_t}$, and the keyframe poses within $\mathcal{W}_t$.
The optimization follows a scheduled iterative strategy.
In the initial phase, we adopt an alternating optimization scheme with two steps.  
First, a deformation probability estimation module infers the posterior probability of non-rigid motion for each Gaussian primitive, denoted as $w_{d,i}^*$ (Sec.~\ref{subsec:prob_estimation}).  
These posterior estimates are treated as pseudo ground truth to supervise the learning of deformation probabilities. Next, we jointly refine all parameters, including the canonical Gaussian primitives, deformation field parameters, and camera poses, by minimizing the following objective:
\begin{equation}
\mathcal{L}_{\text{total}}
=
\mathcal{L}_{\text{ph}}
+
\mathcal{L}_{\text{geo}}
+
\lambda_w\mathcal{L}_w
+
\lambda_{\text{temp}} \mathcal{L}_{\text{temp}}
+
\lambda_{\text{spatial}} \mathcal{L}_{\text{spatial}},
\label{eq:total_loss}
\end{equation}
where $\mathcal{L}_{\text{ph}}$ and $\mathcal{L}_{\text{geo}}$ denote the photometric and geometric losses, respectively.  
The term $\mathcal{L}_w = \sum_{i} \mathrm{BCE}(w_{d,i}, w_{d,i}^*)$ supervises the deformation probabilities using the estimated posteriors.  
To encourage smoothness and consistency, we include a temporal regularization term
$\mathcal{L}_{\text{temp}} = \sum_{i} \lVert w_{d,i}^{t} - w_{d,i}^{t-1} \rVert_2^2$
and a spatial regularization term
$\mathcal{L}_{\text{spatial}} = \sum_{i} \sum_{j \in \mathcal{N}(i)} \lVert w_{d,i} - w_{d,j} \rVert_2^2$,
where $\mathcal{N}(i)$ denotes the neighboring Gaussians of the $i$-th Gaussian.  
These regularization terms improve robustness under noisy or incomplete observations.

Upon reaching the midpoint of the scheduled iterations, the deformation field parameters become stable, and we activate the dynamic deformation field management module (Sec.~\ref{subsec:dynamic_management}).  
Specifically, the number of temporal Gaussian basis functions assigned to each primitive is adaptively adjusted according to the observed deformation complexity.  
This allows the deformation field to expand in deformable regions while remaining compact in stable areas.  
After this structural adaptation, the optimization continues with the same alternating scheme until convergence.

\subsection{Deformation Probability Estimation}
\label{subsec:prob_estimation}

Accurate estimation of the deformation probability $w_{d,i}$ is essential for decoupling camera ego-motion from intrinsic scene deformation.  
Since ground-truth non-rigidity labels are unavailable, we adopt a self-supervised estimation strategy.  
We formulate a probabilistic framework that leverages multi-view observations within the temporal sliding window to estimate the posterior probability of non-rigid motion for each Gaussian primitive.  
In the following, we first formalize this probabilistic framework and subsequently detail the inference of deformation probabilities from image observations.

\subsubsection{Probabilistic Formulation}

To derive the supervision signal $w_{d,i}^*$, we model the motion of the $i$-th Gaussian as a binary classification problem within a Bayesian framework.  
We introduce a latent variable $z_i \in \{R, D\}$ to represent the {rigid} and {deformable} motion modes. The prior probabilities are defined as $P(z_i = R) = \pi_r$ and $P(z_i = D) = \pi_d$, with $\pi_r + \pi_d = 1$.  
The observation likelihood is modeled as a Boltzmann distribution based on photometric inconsistency. For a motion mode $m \in \{R, D\}$, the likelihood is defined as:
\begin{equation}
P(\text{obs} \mid z_i=m) \propto \exp\!\left(-\beta \bar{E}_i^m\right),
\end{equation}
where $\bar{E}_i^m$ denotes the photometric energy of Gaussian $i$ under motion mode $m$, and $\beta = 200$ is the inverse temperature parameter.  
Applying Bayes' rule, the posterior probability that the $i$-th Gaussian follows the deformable motion model is
\begin{equation}
\label{eq:posterior_prob}
P(z_i=D \mid \text{obs}) = 
\frac{\pi_d \exp(-\beta \bar{E}_i^D)}
{\pi_d \exp(-\beta \bar{E}_i^D) + \pi_r \exp(-\beta \bar{E}_i^R)}.
\end{equation}
This posterior defines the pseudo-ground-truth responsibility, denoted as $w_{d,i}^*$.  
For numerical stability and improved gradient flow when supervising $w_{d,i}$, we rewrite it in sigmoid form:
\begin{equation}
\label{eq:posterior_sigmoid}
w_{d,i}^* = \sigma\!\left( 
\log \frac{\pi_d}{\pi_r} 
+ \beta \left( \bar{E}_i^R - \bar{E}_i^D \right) 
\right).
\end{equation}
Here, $\log(\pi_d/\pi_r)$ acts as a prior bias term, while $\beta (\bar{E}_i^R - \bar{E}_i^D)$ represents the data-driven evidence.  
Intuitively, $w_{d,i}^*$ approaches 1 when the deformation model significantly reduces the photometric error compared with the rigid model, i.e., $\bar{E}_i^D \ll \bar{E}_i^R$.  
Therefore, $w_{d,i}^*$ provides a reliable observation-based target for optimizing the learnable parameter $w_{d,i}$.

To apply this framework, we compute the energies $\bar{E}_i^R$ and $\bar{E}_i^D$ for each Gaussian.
The following describes their computation through image observations.

\subsubsection{Dual-Hypothesis Rendering}
To derive the energy terms, we evaluate two motion hypotheses for each keyframe $k \in \mathcal{W}_t$ by rendering the corresponding Gaussian sets under the estimated camera pose $\mathbf{T}_k$. \textit{(i) Rigid Hypothesis ($R$):} This hypothesis enforces a static scene assumption with respect to the canonical space. 
By setting $w_{d,i}=0$, we nullify temporal offsets, projecting the canonical Gaussian map $\mathbf{G}^c$ solely based on the estimated camera pose $\mathbf{T}_k$, i.e., $\hat{\mathbf{I}}_k^R = \mathcal{R}( \mathbf{G}^c, \mathbf{T}_k )$. \textit{(ii) Deformable Hypothesis ($D$):} This hypothesis attributes observation dynamics entirely to the learned deformation field. 
We set $w_{d,i}=1$ to activate the temporal deformation module $\mathcal{D}_{\mathbf{\theta_{t_k}}}$, rendering the deformed primitives as $\hat{\mathbf{I}}_k^D = \mathcal{R}( \mathcal{D}_{\theta_{t_k}}(\mathbf{G}^c, t_k), \mathbf{T}_k )$.
Based on these renderings, we compute the photometric residuals for each hypothesis:
\begin{equation}
\begin{aligned}
    \ell_k^R(\mathbf{u}) &= \mathbf{M}_k^{\text{val}}(\mathbf{u}) \,
    \rho\!\big( \mathbf{I}_k(\mathbf{u}) - \hat{\mathbf{I}}_k^R(\mathbf{u}) \big), \\
    \ell_k^D(\mathbf{u}) &= \mathbf{M}_k^{\text{val}}(\mathbf{u}) \,
    \rho\!\big( \mathbf{I}_k(\mathbf{u}) - \hat{\mathbf{I}}_k^D(\mathbf{u}) \big),
\end{aligned}
\end{equation}
where $\rho(\cdot)$ denotes the $L_1$ robust penalty.  
These residuals are non-negative and are subsequently aggregated to individual Gaussian primitives.  
The validity mask $\mathbf{M}_k^{\text{val}}$ excludes pixels with unreliable measurements.

\subsubsection{Intra-Frame Aggregation}
Given the pixel-wise residuals, we aggregate them to individual Gaussian primitives to obtain element-wise error attribution.
Let $i_{k,j}(\mathbf{u})$ denote the index of the Gaussian primitive contributing to pixel $\mathbf{u}$ at the $j$-th depth-sorted layer, and let $a_{k,j}(\mathbf{u})$ denote its corresponding contribution weight, defined as $a_{k,j}(\mathbf{u}) = \alpha_{k,j} \prod_{l=1}^{j-1} (1 - \alpha_{k,l})$.
They are inherent to the differentiable rasterization process.

We then accumulate the hypothesis-specific energies and effective visibility of the $i$-th Gaussian in keyframe $k$:
\begin{equation}
\begin{aligned}
    E_{i,k}^R &= \sum_{\mathbf{u}} \sum_{j} \mathbb{I}\!\left[ i_{k,j}(\mathbf{u}) = i \right] \, a_{k,j}(\mathbf{u}) \, \ell_k^R(\mathbf{u}), \\
    E_{i,k}^D &= \sum_{\mathbf{u}} \sum_{j} \mathbb{I}\!\left[ i_{k,j}(\mathbf{u}) = i \right] \, a_{k,j}(\mathbf{u}) \, \ell_k^D(\mathbf{u}), \\
    v_{i,k} &= \sum_{\mathbf{u}} \sum_{j} \mathbb{I}\!\left[ i_{k,j}(\mathbf{u}) = i \right] \, a_{k,j}(\mathbf{u}) \, \mathbf{M}_k^{\text{val}}(\mathbf{u}),
\end{aligned}
\end{equation}
where $\mathbb{I}[\cdot]$ is the indicator function.
$E_{i,k}^R$ and $E_{i,k}^D$ quantify the inconsistency of Gaussian $i$ under rigid and deformable assumptions, respectively. 
The visibility term $v_{i,k}$ captures how reliably Gaussian $i$ is observed in keyframe $k$.

\subsubsection{Temporal Window Aggregation}
To improve robustness, we aggregate statistical evidence over a sliding window of keyframes $\mathcal{W}_t = \{k_1, \dots, k_M\}$. 
Direct averaging in the probability space is suboptimal for evidence fusion. 
Instead, we operate in the log-odds (logit) space, where independent evidence can be accumulated additively. 
Specifically, consistent with the posterior definition in Eq.~\eqref{eq:posterior_sigmoid}, the local evidence of Gaussian $i$ in keyframe $k$ is expressed as
\begin{equation}
\label{eq:log_odds}
    \mathcal{L}_{i,k}
    = \log\!\left(\frac{\pi_d}{\pi_r}\right)
    + \beta \big( E_{i,k}^R - E_{i,k}^D \big).
\end{equation}
We then fuse these local estimates to compute the aggregated log-odds $\bar{\mathcal{L}}_i$ via a weighted average over $\mathcal{W}_t$:
\begin{equation}
\label{eq:temporal_log_odds}
    \bar{\mathcal{L}}_i = \frac{\sum_{k \in \mathcal{W}_t} \gamma_{i,k} \, \mathcal{L}_{i,k}}{\sum_{k \in \mathcal{W}_t} \gamma_{i,k} + \varepsilon},
\end{equation}
where $\varepsilon = 1e^{-6}$ is a small constant for numerical stability.
The composite weight $\gamma_{i,k} = \lambda_k \, v_{i,k}$ accounts for both observation reliability and temporal relevance. 
It combines the visibility term $v_{i,k}$ with a temporal decay factor $\lambda_k = \exp(-\eta (t - t_k))$ where $t$ denotes the current timestamp and $t_k$ is the timestamp of keyframe $k$ within $\mathcal{W}_t$. 
The parameter $\eta = 40$ controls the temporal decay rate. Finally, the aggregated evidence is mapped back to the probability space via the sigmoid function:
\begin{equation}
\label{eq:final_dynamic_prob}
    w_{d,i}^* = \sigma(\bar{\mathcal{L}}_i) = \frac{1}{1 + \exp(-\bar{\mathcal{L}}_i)}.
\end{equation}
The resulting $w_{d,i}^*$ serves as a pseudo-ground-truth signal for supervising the learning of deformation probability $w_{d,i}$.

\subsection{Dynamic Deformation Field Management}
\label{subsec:dynamic_management}

As scene deformation evolves, a fixed number of temporal Gaussian basis functions may become insufficient for modeling complex motion or redundant in quasi-static regions. 
To balance representational capacity and computational efficiency, we introduce an adaptive mechanism that dynamically adjusts the number of basis functions $K_i^{\mathcal{A}}$ for each attribute $\mathcal{A} \in \{\boldsymbol{\mu}, \mathbf{s}, \mathbf{q}\}$ of every Gaussian primitive. 
This consists of four operations: densification, merging, pruning, and freezing.

\subsubsection{Densification Strategy}
\label{subsec:densification}

We introduce new temporal basis functions to increase expressivity in regions exhibiting complex deformation. Densification is triggered independently for each attribute under two conditions.

\textbf{Coverage Maintenance.}
We first define the {temporal coverage} to assess whether the evolution of attribute $\mathcal{A}$ is adequately supported at the current timestamp $t_{\mathrm{curr}}$:
\begin{equation}
    C_i^{\mathcal{A}}(t_{\mathrm{curr}}) =
    \sum_{k=1}^{K_i^{\mathcal{A}}}
    \phi\!\left(
        t_{\mathrm{curr}};
        \tau_{i,k}^{\mathcal{A}},
        \sigma_{i,k}^{\mathcal{A}}
    \right).
\end{equation}
If $C_i^{\mathcal{A}}(t_{\mathrm{curr}}) < \delta_{\mathrm{cov}}$, the current basis set provides insufficient support. 
We then insert a new basis function initialized as:
\begin{equation}
\left(
\tau_{i,K_i^{\mathcal{A}}+1}^{\mathcal{A}},
\sigma_{i,K_i^{\mathcal{A}}+1}^{\mathcal{A}},
\omega_{i,K_i^{\mathcal{A}}+1}^{\mathcal{A}}
\right)
=
\left(
t_{\mathrm{curr}},
\sigma_{\mathrm{init}},
0
\right),
\end{equation}
where $\sigma_{\mathrm{init}}$ is set to the mean temporal extent of existing basis functions for attribute $\mathcal{A}$.

\textbf{Error-driven Refinement.}
To capture under-modeled high-frequency deformation, we identify pixels with large photometric error: $\mathcal{P}_{\mathrm{bad}} = \{\mathbf{u} \mid \lVert \hat{\mathbf{I}}_k(\mathbf{u}) - \mathbf{I}_k(\mathbf{u}) \rVert_2 > \tau_{\mathrm{rgb}} \}$.
For each $\mathbf{u} \in \mathcal{P}_{\mathrm{bad}}$, we retrieve the set $\mathcal{N}_{\mathbf{u}}$ of contributing Gaussian primitives. 
To avoid increasing model complexity in rigid regions, we compute a weighted error score specifically for potentially deformable primitives:
\begin{equation}
    E_{\mathrm{def}} =
    \sum_{\mathbf{u} \in \mathcal{P}_{\mathrm{bad}}}
    \sum_{i \in \mathcal{N}_{\mathbf{u}}}
    \alpha_i(\mathbf{u}) \,
    w_{d,i} \,
    \mathbb{I}(w_{d,i} > \tau_{\mathrm{prob}}),
\end{equation}
where $\alpha_i(\mathbf{u})$ denotes the blending weight. 
If $E_{\mathrm{def}}$ exceeds a threshold $\tau_{\mathrm{err}}$, we insert a new basis function at $t_{\mathrm{curr}}$ with  temporal extent $\sigma_{\mathrm{new}} = 0.7 \sigma_{\mathrm{init}}$ and zero initial weight.

\subsubsection{Merging Strategy}
\label{subsec:merging}

To reduce redundancy, we merge similar basis functions within the same attribute. We identify a set of mutually similar basis functions 
$\mathcal{S} \subseteq \{1,\dots,K_i^{\mathcal{A}}\}$ 
such that, for any pair $j,k \in \mathcal{S}$,
\begin{equation}
\left|
\tau_{i,j}^{\mathcal{A}} - \tau_{i,k}^{\mathcal{A}}
\right|
<
\eta_{\mu}
\min\!\left(
\sigma_{i,j}^{\mathcal{A}},
\sigma_{i,k}^{\mathcal{A}}
\right)
\;\text{and}\;
\frac{
\left|
\sigma_{i,j}^{\mathcal{A}} - \sigma_{i,k}^{\mathcal{A}}
\right|
}{
\max\!\left(
\sigma_{i,j}^{\mathcal{A}},
\sigma_{i,k}^{\mathcal{A}}
\right)
}
<
\eta_{\sigma}.
\end{equation}
These functions are fused into a single basis by averaging their parameters and summing their weights:
\begin{equation}
\left(
\tau_{i,\mathrm{new}}^{\mathcal{A}},
\sigma_{i,\mathrm{new}}^{\mathcal{A}},
\omega_{i,\mathrm{new}}^{\mathcal{A}}
\right)
=
\left(
\frac{1}{|\mathcal{S}|}\sum_{k\in\mathcal{S}}\tau_{i,k}^{\mathcal{A}},
\;
\frac{1}{|\mathcal{S}|}\sum_{k\in\mathcal{S}}\sigma_{i,k}^{\mathcal{A}},
\;
\sum_{k\in\mathcal{S}}\omega_{i,k}^{\mathcal{A}}
\right).
\end{equation}
The original basis functions in $\mathcal{S}$ are subsequently removed.

\subsubsection{Pruning and Freezing}
\label{subsec:pruning_freezing}

To maintain a compact deformation field, we identify basis functions with negligible contribution. 
We define the effective activation of the $k$-th basis of attribute $\mathcal{A}$ over a sliding window $\mathcal{W}$ as
\begin{equation}
    u_{i,k}^{\mathcal{A}} =
    \sum_{t \in \mathcal{W}}
    w_{d,i} \,
    \bigl\lVert \omega_{i,k}^{\mathcal{A}} \bigr\rVert \,
    \phi\!\left(
        t;\tau_{i,k}^{\mathcal{A}}, \sigma_{i,k}^{\mathcal{A}}
    \right).
\end{equation}
It is modulated by the deformation probability $w_{d,i}$, so basis functions associated with rigid primitives naturally yield low activation. Then,
a basis function is considered inactive if its relative effective activation satisfies $
\frac{u_{i,k}^{\mathcal{A}}}
{\sum_{j=1}^{K_i^{\mathcal{A}}} u_{i,j}^{\mathcal{A}}}
<
\delta_{\mathrm{act}}.
$
For inactive bases, we distinguish two cases according to temporal relevance.
If $
\tau_{i,k}^{\mathcal{A}} + 3\sigma_{i,k}^{\mathcal{A}} < t_{\mathrm{curr}},
$, the basis is considered temporally expired. 
Its parameters are frozen to preserve historical deformation states while excluding it from further optimization.
Otherwise, if the basis remains temporally relevant but exhibits negligible activation, it is regarded as redundant and removed.

\subsection{System Initialization}
\label{subsec:initialization}

The initialization process bootstraps the canonical map and the deformation field using an initial window of frames. This procedure follows a four-step sequence:

\textbf{(i) Canonical Map Initialization.}
Consistent with standard protocols in GS-based SLAM \cite{matsuki2024gaussian,wang2024endogslam}, we initialize the canonical map using the first frame of the sequence. 
The global coordinate system is defined by setting the initial camera pose to the identity transformation, $\mathbf{T}_0 = \mathbf{I}$. 
Using the depth prior $\hat{\mathbf{D}}_0$ from the measurement preprocessing module, we back-project image pixels into 3D space to instantiate the canonical Gaussian primitives. 
We then optimize the canonical parameters by minimizing the photometric and depth losses for the first frame, establishing a reliable base geometry.

\textbf{(ii) Pose-Only Optimization.}
For the remaining frames in the initial window $\mathcal{W}_t$, we perform pose-only optimization to align the camera views with the initialized map:
$
    \min_{ \{ \mathbf{T}_{k} \} } 
    \sum_{k \in \mathcal{W}_t}
    \sum_{\mathbf{u}}
    \left( 
        \mathcal{L}_{\text{ph}}(\mathbf{u}, \mathbf{T}_k) 
        + 
        \lambda_{\text{geo}} \mathcal{L}_{\text{geo}}(\mathbf{u}, \mathbf{T}_k)
    \right).
$

\textbf{(iii) Deformation Attribute Initialization.}
After pose alignment, we initialize the deformation-related parameters. 
First, the deformation probabilities $w_{d,i}$ are set to $0.6$, representing a weak prior that the scene may be deformable. 
Second, we initialize the temporal basis functions. 
Each attribute $\mathcal{A} \in \{\boldsymbol{\mu}, \mathbf{s}, \mathbf{q}\}$ is assigned $K^{\mathcal{A}}$ basis functions, with temporal centers $\tau$ uniformly distributed over the window. 
Let $\Delta t$ denote the interval between adjacent centers; the temporal extent is set to $\sigma = 0.7 \Delta t$. 
This choice ensures sufficient overlap between neighboring bases for smooth temporal transitions. 
All deformation coefficients are initialized to zero.

\textbf{(iv) Global Deformable Bundle Adjustment.}
Finally, we perform a global deformable bundle adjustment over the window, as described in Sec.~\ref{subsec:global_ba}, to establish a consistent foundation for subsequent deformable tracking and mapping.
\section{Experiments}
\subsection{Experimental Setup}

\begin{table*}[t]
\centering
\setlength{\tabcolsep}{1.2pt}
\renewcommand{\arraystretch}{1.15}
\caption{Quantitative comparison of camera localization accuracy (ATE) on the StereoMIS dataset. Values are reported as RMSE and Standard Deviation (S.D.) in millimeters (mm). ``Clip'' denotes evaluation on a continuous 1,000-frame segment. $\times$ indicates tracking failure. $\dagger$ denotes that the average is computed over successfully tracked sequences only.}
\label{tab:camera_localization_stereomis}

\begin{tabular}{l c c c c c c c c c c c c c c c c c c c c}
\toprule
\multirow{3}{*}{\textbf{Methods}}
& \multicolumn{4}{c}{\textbf{P2-2}}
& \multicolumn{4}{c}{\textbf{P2-3}}
& \multicolumn{4}{c}{\textbf{P2-4}}
& \multicolumn{4}{c}{\textbf{P2-5}}
& \multicolumn{4}{c}{\textbf{Avg.}} \\
\cmidrule(lr){2-5}
\cmidrule(lr){6-9}
\cmidrule(lr){10-13}
\cmidrule(lr){14-17}
\cmidrule(lr){18-21}

& \multicolumn{2}{c}{Clip} & \multicolumn{2}{c}{Full}
& \multicolumn{2}{c}{Clip} & \multicolumn{2}{c}{Full}
& \multicolumn{2}{c}{Clip} & \multicolumn{2}{c}{Full}
& \multicolumn{2}{c}{Clip} & \multicolumn{2}{c}{Full}
& \multicolumn{2}{c}{Clip} & \multicolumn{2}{c}{Full} \\

& RMSE & S.D. & RMSE & S.D.
& RMSE & S.D. & RMSE & S.D.
& RMSE & S.D. & RMSE & S.D.
& RMSE & S.D. & RMSE & S.D.
& RMSE & S.D. & RMSE & S.D. \\
\midrule

\multicolumn{21}{l}{\textcolor{gray}{\textit{\textbf{Traditional Monocular Non-rigid SLAM}}}} \\

DefSLAM
& 48.45 & 19.22 & 28.96 & 21.99 
& 0.026 & 0.018 & 3.93 & 3.75
& 33.73 & 16.34 & 54.54 & 23.64
& 28.46 & 13.19 & 60.85 & 17.97
& 27.67 & 12.19 & 37.07 & 16.84\\

NR-SLAM
& 38.62 & 16.85 & 28.22 & 20.60
& 0.026 & 0.018 & 2.69 & 2.43
& 23.97 & 10.70 & 40.24 & 18.87
& $\times$ & $\times$ & $\times$ & $\times$ 
& 20.87$^{\dagger}$ & 9.19$^{\dagger}$ & 23.72$^{\dagger}$ & 13.97$^{\dagger}$\\

\midrule
\multicolumn{21}{l}{\textcolor{gray}{\textit{\textbf{General-Purpose GS-based SLAM}}}} \\

MonoGS
& 47.58 & 22.11 & 66.29 & 22.81
& 0.003 & 0.002 & 4.48 & 1.92
& 51.37 & 21.91 & 75.49 & 25.03
& 59.98 & 22.08 & 70.57 & 24.63
& 39.73 & 16.53 & 54.21 & 18.60 \\

S3PO
& 19.75 & 10.66 & 28.21 & 11.55
& 0.003 & 0.002 & 3.48 & 1.77
& 18.09 & 7.58 & 26.12 & 13.29
& 16.75 & 5.95 & 26.71 & 13.73
& 13.65 & 6.05 & 21.13 & 10.09 \\

4DTAM
& 21.34 & 10.59 & 34.11 & 17.79
& 0.004 & 0.002 & 6.28 & 2.35
& 27.41 & 12.92 & 36.24 & 16.92
& 14.28 & 9.92 & 29.70 & 14.07
& 15.76 & 8.36 & 26.58 & 12.78 \\

\midrule
\multicolumn{21}{l}{\textcolor{gray}{\textit{\textbf{Endoscopy-Specific Differentiable Rendering SLAM}}}} \\

DDS-SLAM
& 19.40 & 9.63 & 17.76 & 12.19
& 0.026 & 0.018 & 3.70 & 3.38
& 21.66 & 8.63 & 37.82& 17.11
& 20.21 & 8.09 & 52.42& 24.95
& 15.32 & 6.59 & 27.93& 14.41\\

EndoGSLAM
& 39.68 & 17.80 & $\times$ &  $\times$ 
& 0.001 & 0.001 & $\times$ &  $\times$ 
& 34.95 & 16.58 & $\times$ &  $\times$ 
& 33.84 & 14.28 & $\times$ &  $\times$ 
& 27.12 & 12.17 & $\times$ &  $\times$ \\

Endo-2DTAM
& 36.36 & 18.52 & $\times$ &  $\times$ 
& 0.013 & 0.006 & $\times$ &  $\times$ 
& 31.06 & 14.00 & $\times$ &  $\times$ 
& 33.35 & 12.63 & $\times$ &  $\times$ 
& 25.20 & 11.29 & $\times$ &  $\times$ \\

\midrule
\rowcolor{gray!10}
NRGS-SLAM (Ours)
& \textbf{10.24} & \textbf{4.90} & \textbf{16.06}  & \textbf{7.89}
& 0.003 & 0.002 & \textbf{2.17} & \textbf{1.04}
& \textbf{9.45} & \textbf{4.63} & \textbf{14.22} & \textbf{6.28} 
& \textbf{7.41} & \textbf{3.03} & \textbf{15.85}  & \textbf{8.74}
& \textbf{6.78} & \textbf{3.14} & \textbf{12.08} & \textbf{5.99} \\

\bottomrule
\end{tabular}
\end{table*}

\subsubsection{Datasets}
We evaluate  NRGS-SLAM on three public endoscopic datasets: StereoMIS~\cite{hayoz2023learning}, Hamlyn~\cite{mountney2010three,stoyanov2005soft,stoyanov2010real,pratt2010dynamic}, and C3VDv2~\cite{golhar2025c3vdv2}. These datasets are widely utilized in prior research and contain diverse types of deformations. Although the EndoMapper dataset~\cite{azagra2023endomapper} is also a prominent benchmark in the field, it was not accessible during this study. Detailed descriptions are provided below:

\begin{itemize}
    \item \textbf{StereoMIS Dataset} contains in-vivo endoscopic videos recorded with the da Vinci Xi surgical system. The public release includes 11 sequences from three porcine subjects, capturing tissue deformation caused by respiration and instrument manipulation. Ground-truth camera trajectories derived from robot kinematics are provided for quantitative evaluation of localization accuracy.

    \item \textbf{Hamlyn Dataset} is collected  by the Hamlyn Centre at Imperial College London and contains a diverse collection of laparoscopic and endoscopic videos. It captures deformations induced by respiration, heartbeat, and instrument-tissue interactions. In this work, we utilize the rectified version of the dataset provided by~\cite{recasens2021endo}.

    \item \textbf{C3VDv2 Dataset} offers enhanced scale and realism compared to its predecessor~\cite{bobrow2023colonoscopy}. We evaluate our method on the colon deformation subset, where non-rigid motion is externally induced via balloon pressurization, accompanied by challenging environmental conditions such as occlusions by fecal matter. Furthermore, it provides ground-truth camera poses for localization accuracy assessment.
\end{itemize}

\subsubsection{Evaluation Metrics}
To evaluate localization performance, we adopt  the widely used Absolute Trajectory Error (ATE)~\cite{zhang2018tutorial}, which measures the discrepancy between estimated and ground-truth camera trajectories. We report both the Root Mean Square Error (RMSE) and the Standard Deviation (SD) of ATE. All trajectory errors are reported in millimeters. To assess mapping performance, following prior works \cite{li2025pg,wu20244d}, we evaluate rendered image fidelity using Peak Signal-to-Noise Ratio (PSNR), Structural Similarity Index (SSIM)\cite{wang2004image}, and Learned Perceptual Image Patch Similarity (LPIPS)\cite{lpips}.

\subsubsection{Implementation Details}

All experiments were conducted on an NVIDIA RTX 4090 GPU.
In the \textbf{measurement preprocessing}, the hyperparameters of \cite{xiao2025spatialtrackerv2} are set as follows: grid size is 6, buffer size is 3. The validity mask threshold is 0.2, and the transparency threshold is 0.1.
In \textbf{deformable tracking}, the exponential annealing schedule for the geometric loss uses $\lambda_g^0 = 1.0$, $\lambda_g^{\min} = 0.01$, and $\tau_g = 10$. The deformation filtering threshold in the first stage is $\tau_{\mathrm{def}} = 0.5$, and the second stage runs for 40 iterations. The learning rates are set to $4\times10^{-4}$ for camera rotation  and $1\times10^{-3}$ for camera translation. In per-frame deformation estimation, the deformation update gating threshold is $\epsilon_{\mathrm{def}} = 0.5$, with regularization weights $\lambda_{\mathrm{reg}} = 0.01$ and $\lambda_{\mathrm{tem}} = 0.01$, a learning rate of $5\times10^{-4}$, and 40  iterations.
In the \textbf{deformable mapping}, the sliding window size is set to 7, and keyframe poses optimization runs for 20 iterations. In global deformable bundle adjustment, $\lambda_w = 1$, $\lambda_{\mathrm{temp}} = 0.01$, and $\lambda_{\mathrm{spatial}} = 0.01$. The initial learning rates are set to $1.6\times10^{-4}$ for Gaussian means, scale and rotation, $2\times10^{-4}$ for camera rotation, and $5\times10^{-4}$ for camera translation. The initial learning rate of the deformation field is $1.6\times10^{-4}$.
In dynamic deformation field management, the coverage maintenance threshold is $\delta_{\mathrm{cov}} = 0.5$, the RGB threshold is $\tau_{\mathrm{rgb}} = 5\times10^{-2}$, and the probability threshold is $\tau_{\mathrm{prob}} = 0.5$. In the merging strategy, $\eta_{\mu} = 0.5$ and $\eta_{\sigma} = 1\times10^{-1}$. In pruning and freezing, the activity threshold is $\delta_{\mathrm{act}} = 5\times10^{-2}$.

\begin{table*}[t]
\centering
\setlength{\tabcolsep}{3.5pt}
\renewcommand{\arraystretch}{1.15}
\caption{Quantitative comparison of camera localization accuracy (ATE) on the C3VDv2 dataset. Values are reported as RMSE and Standard Deviation (S.D.) in millimeters (mm). $\times$ indicates tracking failure. $\dagger$ denotes that the average is computed over successfully tracked sequences only.}
\label{tab:camera_localization_c3vdv2}

\begin{tabular}{l c c c c c c c c c c}
\toprule
\multirow{2}{*}{\textbf{Methods}}
& \multicolumn{2}{c}{\textbf{c1\_descending\_t4\_v4}}
& \multicolumn{2}{c}{\textbf{c1\_sigmoid1\_t4\_v4}}
& \multicolumn{2}{c}{\textbf{c1\_sigmoid2\_t4\_v4}}
& \multicolumn{2}{c}{\textbf{c2\_transverse1\_t1\_v4}}
& \multicolumn{2}{c}{\textbf{Avg.}} \\
\cmidrule(lr){2-3}
\cmidrule(lr){4-5}
\cmidrule(lr){6-7}
\cmidrule(lr){8-9}
\cmidrule(lr){10-11}

& RMSE & S.D.
& RMSE & S.D.
& RMSE & S.D.
& RMSE & S.D.
& RMSE & S.D. \\
\midrule

\multicolumn{11}{l}{\textcolor{gray}{\textit{\textbf{Traditional Monocular  non-rigid  SLAM}}}} \\

DefSLAM
& 21.35 & 11.27
& 10.59 & 4.31
& 10.35 & 3.71
& 13.49 & 5.99
& 13.94 & 6.32 \\

NR-SLAM
& 16.84 & 10.74
& $\times$ & $\times$
& \textbf{6.09} & \textbf{2.87}
& 10.93 & 6.29
& 11.29$^{\dagger}$ & 6.63$^{\dagger}$ \\

\midrule

\multicolumn{11}{l}{\textcolor{gray}{\textit{\textbf{General-Purpose GS-based SLAM}}}} \\

MonoGS
& 34.56 & 16.70
& 27.44 & 11.80
& 29.07 & 15.95
& 35.65 & 18.25
& 31.67 & 15.68 \\

S3PO
& 24.31 & 9.11
& 18.65 & 7.57
& 29.04 & 13.72
& 30.75 & 15.51
& 25.69 & 11.48 \\

4DTAM
& 13.35 & 5.75
& 12.70 & 5.26
& 19.54 & 8.28
& 15.84 & 6.82
& 15.36 & 6.53 \\

\midrule

\multicolumn{11}{l}{\textcolor{gray}{\textit{\textbf{Endoscopy-Specific Differentiable Rendering SLAM}}}} \\

DDS-SLAM
& 11.57 & 4.94
& 10.63 & 3.74
& 10.91 & 4.34
& 12.99 & 3.64
& 11.52 & 4.16 \\

EndoGSLAM
& 22.60 & 9.64
& 16.99 & 7.87
& 17.01 & 7.33
& 47.28 & 20.38
& 25.97 & 11.81 \\

Endo-2DTAM
& 30.33 & 9.83
& 17.30 & 7.62
& 30.57 & 13.69
& 49.58 & 18.67
& 31.95 & 12.45 \\

\midrule

\rowcolor{gray!10}
NRGS-SLAM (Ours)
& \textbf{6.81} & \textbf{3.27}
& \textbf{7.96} & \textbf{3.66}
& {7.26} & {3.87} 
& \textbf{10.47} & \textbf{4.88}
& \textbf{8.13} & \textbf{3.92}\\

\bottomrule
\end{tabular}
\end{table*}
\begin{figure}[t]
\centerline{\includegraphics[width=0.5\textwidth]{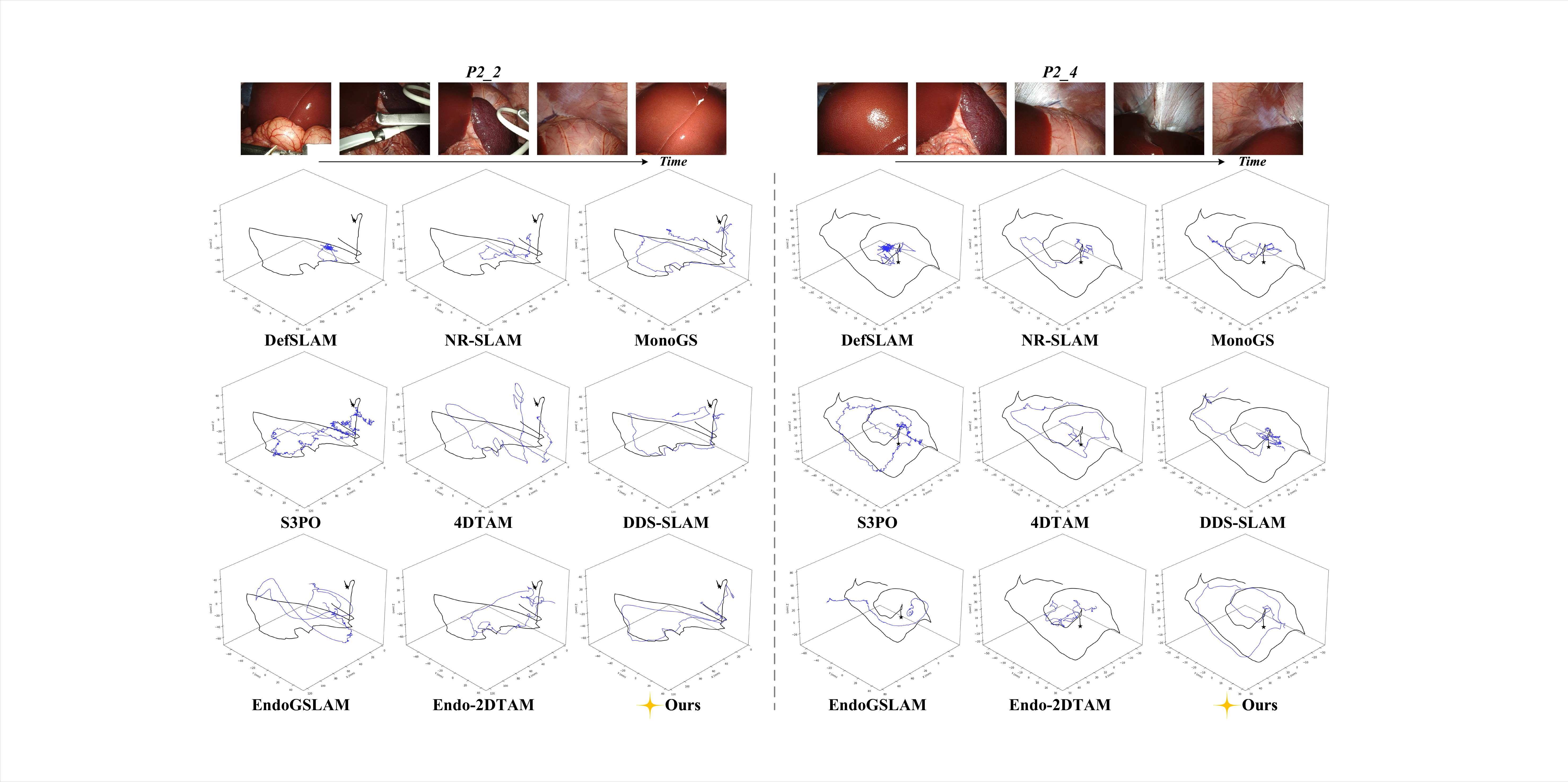}}
\caption{Qualitative comparison of camera trajectory estimation on the StereoMIS dataset. The top row presents representative image frames in temporal order. The subsequent rows show 3D trajectory comparisons of different methods. Estimated trajectories are shown in blue and ground truth in black. The star marker denotes the starting point.}
    \label{fig:traj stereomisv2}
\end{figure}

\subsection{Comparison with State-of-the-art Approaches}
\subsubsection{Methods for Comparison}

To validate the effectiveness of NRGS-SLAM, we conduct comprehensive comparisons against representative state-of-the-art approaches discussed in Sec.~\ref{sec:related work}. These baselines are categorized into three groups based on their methodological paradigms and application domains:

\begin{itemize}
    \item \textbf{Traditional Monocular  Non-rigid  SLAM.} This group includes classical approaches based on traditional scene representations, such as points or meshes. We select DefSLAM~\cite{lamarca2020defslam} and NR-SLAM~\cite{rodriguez2024nr} as representative state-of-the-art monocular non-rigid SLAM methods.
    
    \item \textbf{General-Purpose GS-based SLAM.} This group includes GS-based SLAM methods developed for general scenes.We include them due to their methodological relevance. MonoGS\cite{matsuki2024gaussian} is a representative GS-based system and supports multiple sensor modalities. In this paper, we configure it in monocular mode. S3PO~\cite{li2025pg} is an RGB-only system that, similar to our approach, leverages geometric priors to compensate for the lack of depth input. 4DTAM\cite{matsuki20254dtam} combines 2D Gaussian Splatting with deformation fields and supports non-rigid mapping.

    \item \textbf{Endoscopy-Specific Differentiable Rendering SLAM.} This group consists of recent methods tailored for endoscopic environments using implicit neural representations or Gaussian Splatting. DDS-SLAM~\cite{shan2024dds} characterizes deformable endoscopic scenes using neural implicit representations coupled with an MLP-based deformation network. EndoGSLAM~\cite{wang2024endogslam} and Endo-2DTAM~\cite{11128637} recent the latest Gaussian Splatting-based approaches for endoscopy, primarily designed for static scenes. 
\end{itemize}

\noindent\textit{Note on Input Modality:} Several baselines, including 4DTAM, DDS-SLAM, EndoGSLAM, and Endo-2DTAM, natively require RGB-D input, which is incompatible with our monocular setup. For fair comparison, we provide these methods with depth maps generated via offline inference using the same geometric foundation model~\cite{xiao2025spatialtrackerv2} adopted in our system, thereby constructing pseudo RGB-D inputs. In addition, we remove the original semantic mask requirement of DDS-SLAM.

\subsubsection{Camera Localization}
\begin{figure*}[th]
\centerline{\includegraphics[width=0.9\textwidth]{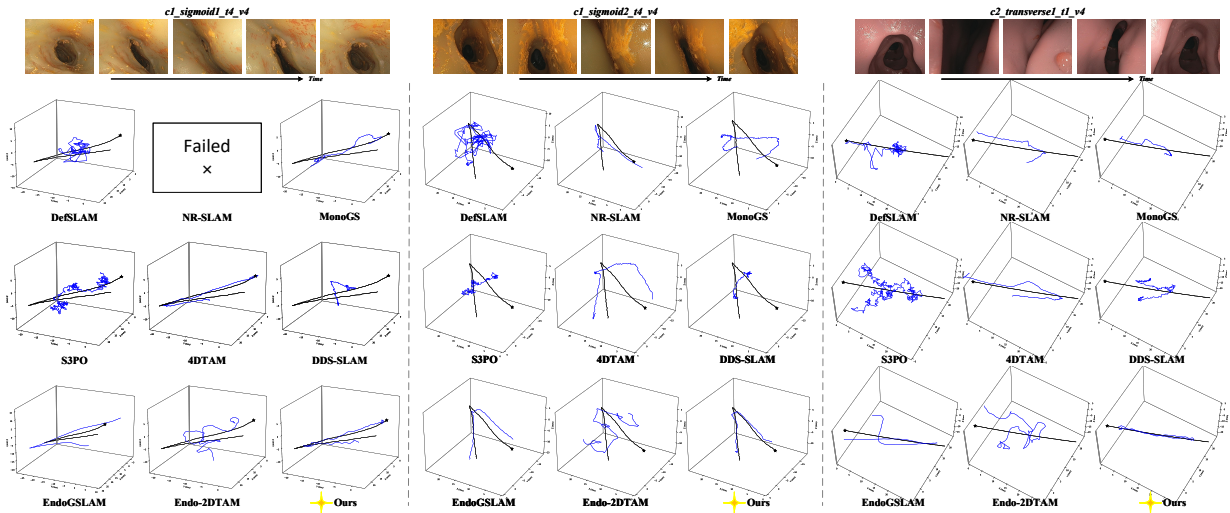}}
\caption{Qualitative comparison of camera trajectory estimation on the C3VDv2 dataset~\cite{hayoz2023learning}. The top row presents representative image frames in temporal order. The subsequent rows show 3D trajectory comparisons of different methods. Estimated trajectories are shown in blue and ground truth in black. The star marker denotes the starting point.}
    \vspace{-15pt}
    \label{fig:traj c3vdv2}
\end{figure*}

We evaluate the camera localization accuracy on the StereoMIS and C3VDv2 datasets. Since the Hamlyn dataset lacks ground truth camera poses, quantitative evaluation is omitted for it. The quantitative results are reported in Table~\ref{tab:camera_localization_stereomis} and Table~\ref{tab:camera_localization_c3vdv2}, with qualitative visualizations provided in Fig.~\ref{fig:traj stereomisv2} and Fig.~\ref{fig:traj c3vdv2}. Overall, NRGS-SLAM consistently outperforms state-of-the-art methods, demonstrating superior robustness and accuracy in challenging deformable endoscopic environments. Detailed analysis is provided below.

\begin{table*}[th]
\centering
\setlength{\tabcolsep}{4.5pt}
\renewcommand{\arraystretch}{1.15}
\caption{Quantitative comparison of rendering quality on the StereoMIS dataset
under Clip (1000 frames) and Full sequence settings.
PSNR (dB) and SSIM are higher-is-better, while LPIPS is lower-is-better.}
\label{tab:rendering_stereomis}

\begin{tabular}{l c c c c c c c c c c c c c c c}
\toprule
\multirow{2}{*}{\textbf{Methods}}
& \multicolumn{3}{c}{\textbf{P2-2 (Clip)}}
& \multicolumn{3}{c}{\textbf{P2-3 (Clip)}}
& \multicolumn{3}{c}{\textbf{P2-4 (Clip)}}
& \multicolumn{3}{c}{\textbf{P2-5 (Clip)}}
& \multicolumn{3}{c}{\textbf{Avg. (Clip)}} \\
\cmidrule(lr){2-4}
\cmidrule(lr){5-7}
\cmidrule(lr){8-10}
\cmidrule(lr){11-13}
\cmidrule(lr){14-16}

& PSNR & SSIM & LPIPS
& PSNR & SSIM & LPIPS
& PSNR & SSIM & LPIPS
& PSNR & SSIM & LPIPS
& PSNR & SSIM & LPIPS \\
\midrule

\multicolumn{16}{l}{\textcolor{gray}{\textit{\textbf{General-Purpose GS-based SLAM}}}} \\

MonoGS
& 16.59 & 0.620 & 0.671
& 23.09 & 0.660 & 0.321
& 16.99 & 0.478 & 0.565
& 15.40 & 0.382 & 0.693
& 18.02 & 0.535 & 0.563 \\

S3PO
& 15.84 & 0.659 & 0.713
& 23.21 & 0.591 & 0.399
& 19.71 & 0.572 & 0.729
& 20.05 & 0.522 & 0.749
& 19.70 & 0.586 & 0.648 \\

4DTAM
& 17.78 & 0.652 & 0.569
& 23.39 & 0.649 & 0.252
& 21.79 & 0.582 & 0.471
& 19.78 & 0.577 & 0.597
& 20.68 & 0.615 & 0.472 \\

\midrule
\multicolumn{16}{l}{\textcolor{gray}{\textit{\textbf{Endoscopy-Specific Differentiable Rendering SLAM}}}} \\

DDS-SLAM 
& 20.64 & 0.586 & 0.613
& 23.54 & 0.480 & 0.537 
& 21.30 & 0.561 & 0.654
& 21.44 & 0.538 & 0.655
& 21.73 & 0.541 & 0.615\\

EndoGSLAM
& 14.72 & 0.592 & 0.511
& 23.85 & 0.475 & 0.332
& 16.95 & 0.431 & 0.448
& 21.38 & 0.503 & \textbf{0.334}
& 19.23 & 0.500 & 0.406 \\

Endo-2DTAM
& 11.87 & 0.527 & 0.665
& 18.32 & 0.463 & 0.638
& 16.06 & 0.503 & 0.572
& 14.06 & 0.434 & 0.622
& 15.08 & 0.482 & 0.624 \\

\midrule
\rowcolor{gray!10}
NRGS-SLAM (Ours)
& \textbf{26.23} & \textbf{0.768} & \textbf{0.456}
& \textbf{26.71} & \textbf{0.720} & \textbf{0.132}
& \textbf{27.32} & \textbf{0.609} & \textbf{0.395}
& \textbf{25.52} & \textbf{0.625} & {0.490}
& \textbf{26.45} & \textbf{0.681} & \textbf{0.368} \\

\bottomrule
\end{tabular}

\vspace{4pt}

\begin{tabular}{l c c c c c c c c c c c c c c c}
\toprule
\multirow{2}{*}{\textbf{Methods}}
& \multicolumn{3}{c}{\textbf{P2-2 (Full)}}
& \multicolumn{3}{c}{\textbf{P2-3 (Full)}}
& \multicolumn{3}{c}{\textbf{P2-4 (Full)}}
& \multicolumn{3}{c}{\textbf{P2-5 (Full)}}
& \multicolumn{3}{c}{\textbf{Avg. (Full)}} \\
\cmidrule(lr){2-4}
\cmidrule(lr){5-7}
\cmidrule(lr){8-10}
\cmidrule(lr){11-13}
\cmidrule(lr){14-16}

& PSNR & SSIM & LPIPS
& PSNR & SSIM & LPIPS
& PSNR & SSIM & LPIPS
& PSNR & SSIM & LPIPS
& PSNR & SSIM & LPIPS \\
\midrule

\multicolumn{16}{l}{\textcolor{gray}{\textit{\textbf{General-Purpose GS-based SLAM}}}} \\

MonoGS
& 13.88 & 0.479 & 0.770
& 20.21 & 0.560 & 0.436
& 14.83 & 0.430 & 0.671
& 13.51 & 0.358 & 0.756
& 15.61 & 0.457 & 0.658 \\

S3PO
& 14.08 & 0.576 & 0.784
& 21.16 & 0.470 & 0.599
& 17.63 & 0.476 & 0.767
& 18.25 & 0.460 & 0.786
& 17.78 & 0.496 & 0.734 \\

4DTAM
& 16.04 & 0.497 & 0.591
& 21.67 & 0.561 & 0.461
& 20.95 & 0.494 & \textbf{0.530}
& 17.22 & 0.457 & 0.610
& 18.97 & 0.502 & 0.548 \\

\midrule
\multicolumn{16}{l}{\textcolor{gray}{\textit{\textbf{Endoscopy-Specific Differentiable Rendering SLAM}}}} \\

DDS-SLAM 
& 19.92	& 0.509 & 0.682 
& 22.27	& 0.462 & 0.559
& 21.22 & 0.526 & 0.680
& 20.89 & 0.526 & 0.697
& 21.08 & 0.506 & 0.655\\

EndoGSLAM
& $\times$ & $\times$ & $\times$
& $\times$ & $\times$ & $\times$
& $\times$ & $\times$ & $\times$
& $\times$ & $\times$ & $\times$
& $\times$ & $\times$ & $\times$ \\

Endo-2DTAM
& $\times$ & $\times$ & $\times$
& $\times$ & $\times$ & $\times$
& $\times$ & $\times$ & $\times$
& $\times$ & $\times$ & $\times$
& $\times$ & $\times$ & $\times$ \\

\midrule
\rowcolor{gray!10}
NRGS-SLAM (Ours)
& \textbf{20.84} & \textbf{0.626} & \textbf{0.540}
& \textbf{24.67} & \textbf{0.652} & \textbf{0.333}
& \textbf{24.39} & \textbf{0.560} & {0.568}
& \textbf{22.89} & \textbf{0.541} & \textbf{0.569}
& \textbf{23.20} & \textbf{0.595} & \textbf{0.503} \\

\bottomrule
\end{tabular}
\vspace{-10pt}
\end{table*}

\textbf{Evaluation on StereoMIS Dataset:}
We evaluate four sequences: P2-2 (4,278 frames), P2-3 (4,223 frames), P2-4 (4,281 frames), and P2-5 (5,368 frames). P2-3 is characterized by tissue deformation with a relatively static camera, while the remaining sequences exhibit large-scale camera motion coupled with scene deformation. Some endoscopy-specific methods (EndoGSLAM and Endo-2DTAM) fail to complete tracking on full-length sequences due to the excessive sequence length and complex dynamics. To enable fair comparison, we additionally evaluate all methods on continuous 1,000-frame segments, denoted as the \textit{Clip} mode.

As shown in Table~\ref{tab:camera_localization_stereomis}, among traditional monocular non-rigid SLAM, NR-SLAM generally outperforms DefSLAM on most sequences; however, it fails on P2-5. General-purpose GS-based methods, such as MonoGS, suffer from significant drift, yielding an average RMSE exceeding 39mm on clips and 54mm on full sequence. While S3PO reduces these errors by leveraging geometric priors, it still exhibits substantial tracking errors. Similarly,  EndoGSLAM and Endo-2DTAM demonstrate improved performance on short clips compared to MonoGS but fail to maintain stability over long durations. Fundamentally, these methods struggle due to their reliance on static assumptions and their inability to handle scene deformations. DDS-SLAM and 4DTAM improve localization accuracy by  modeling non-rigid deformation. Nevertheless, they still exhibit drift, likely due to the ambiguity between camera ego-motion and scene deformation. NRGS-SLAM introduces explicit mechanisms to effectively mitigate this coupling and achieves the lowest localization error among all compared methods, corresponding to a relative reduction of 50.3\% in Clip mode and 42.8\% on full sequences compared with the second-best method. These results are consistent with the qualitative trajectory comparisons in Fig.~\ref{fig:traj stereomisv2}.

\begin{table*}[t]
\centering
\setlength{\tabcolsep}{4pt}
\renewcommand{\arraystretch}{1.15}

\caption{Quantitative comparison of rendering quality on the Hamlyn dataset.}
\label{tab:rendering_hamlyn}

\begin{tabular}{l c c c c c c c c c c c c}
\toprule
\multirow{2}{*}{\textbf{Methods}}
& \multicolumn{3}{c}{\textbf{rectified18}}
& \multicolumn{3}{c}{\textbf{rectified19}}
& \multicolumn{3}{c}{\textbf{rectified23}}
& \multicolumn{3}{c}{\textbf{Avg.}} \\
\cmidrule(lr){2-4}
\cmidrule(lr){5-7}
\cmidrule(lr){8-10}
\cmidrule(lr){11-13}

& PSNR$\uparrow$ & SSIM$\uparrow$ & LPIPS$\downarrow$
& PSNR$\uparrow$ & SSIM$\uparrow$ & LPIPS$\downarrow$
& PSNR$\uparrow$ & SSIM$\uparrow$ & LPIPS$\downarrow$
& PSNR$\uparrow$ & SSIM$\uparrow$ & LPIPS$\downarrow$ \\
\midrule

\multicolumn{13}{l}{\textcolor{gray}{\textit{\textbf{General-Purpose GS-based SLAM}}}} \\

MonoGS
& \textbf{25.69} & 0.848 & 0.457
& 29.30 & 0.889 & {0.455}
& 23.01 & 0.723 & 0.489
& {26.00} & 0.820 & 0.467 \\

S3PO
& 17.93 & 0.751 & 0.465
& 26.56 & 0.880 & 0.496
& 18.90 & 0.678 & 0.489
& 21.13 & 0.770 & 0.484 \\

4DTAM
& 23.04 & 0.863 & 0.428
& 29.07 & 0.901 & 0.470
& {23.04} & 0.747 & 0.458
& 25.05 & 0.837 & 0.452 \\

\midrule

\multicolumn{13}{l}{\textcolor{gray}{\textit{\textbf{Endoscopy-Specific Differentiable Rendering SLAM}}}} \\

DDS-SLAM
& 10.95 & 0.322 & 0.745
& 10.97 & 0.379 & 0.742
& 11.10 & 0.300 & 0.721
& 11.01 & 0.334 & 0.736 \\

EndoGSLAM
& 16.86 & 0.568 & 0.565
& 15.85 & 0.524 & 0.594
& 12.19 & 0.367 & 0.605
& 14.97 & 0.486 & 0.588 \\

Endo-2DTAM
& 17.49 & 0.653 & 0.611
& 19.63 & 0.711 & 0.572
& 16.49 & 0.593 & 0.588
& 17.87 & 0.652 & 0.590 \\

\midrule
\rowcolor{gray!10}
NRGS-SLAM (Ours)
& {24.58} & \textbf{0.880} & \textbf{0.405}
& \textbf{30.58} & \textbf{0.925} &  \textbf{0.427}
& \textbf{24.16} & \textbf{0.775} & \textbf{0.452}
&  \textbf{26.44} & \textbf{0.860} & \textbf{0.428} \\

\bottomrule
\end{tabular}
\end{table*}
\begin{table*}[t]
\centering
\setlength{\tabcolsep}{2.5pt}
\renewcommand{\arraystretch}{1.15}

\caption{Quantitative comparison of rendering quality on the C3VDv2 dataset.}
\label{tab:rendering_c3vdv2}

\begin{tabular}{l c c c c c c c c c c c c c c c}
\toprule
\multirow{2}{*}{\textbf{Methods}}
& \multicolumn{3}{c}{\textbf{c1\_descending\_t4\_v4}}
& \multicolumn{3}{c}{\textbf{c1\_sigmoid1\_t4\_v4}}
& \multicolumn{3}{c}{\textbf{c1\_sigmoid2\_t4\_v4}}
& \multicolumn{3}{c}{\textbf{c2\_transverse1\_t1\_v4}}
& \multicolumn{3}{c}{\textbf{Avg.}} \\
\cmidrule(lr){2-4}
\cmidrule(lr){5-7}
\cmidrule(lr){8-10}
\cmidrule(lr){11-13}
\cmidrule(lr){14-16}

& PSNR$\uparrow$ & SSIM$\uparrow$ & LPIPS$\downarrow$
& PSNR$\uparrow$ & SSIM$\uparrow$ & LPIPS$\downarrow$
& PSNR$\uparrow$ & SSIM$\uparrow$ & LPIPS$\downarrow$
& PSNR$\uparrow$ & SSIM$\uparrow$ & LPIPS$\downarrow$
& PSNR$\uparrow$ & SSIM$\uparrow$ & LPIPS$\downarrow$ \\
\midrule

\multicolumn{16}{l}{\textcolor{gray}{\textit{\textbf{General-Purpose GS-based SLAM}}}} \\

MonoGS
& 17.04 & 0.406 & 0.592
& 16.61 & 0.510 & 0.720
& 17.20 & 0.640 & 0.761
& 15.14 & 0.538 & 0.659
& 16.50 & 0.523 & 0.683 \\

S3PO
& 15.75 & 0.572 & 0.790
& 19.11 & 0.690 & 0.729
& 17.76 & 0.600 & 0.768
& 18.93 & 0.746 & 0.638
& 17.89 & 0.652 & 0.731 \\

4DTAM
& 22.61 & 0.618 & 0.673
& 26.48 & 0.728 & 0.647
& 24.57 & 0.667 & 0.688
& 22.18 & 0.777 & 0.580
& 23.96 & 0.698 & 0.647 \\

\midrule

\multicolumn{16}{l}{\textcolor{gray}{\textit{\textbf{Endoscopy-Specific Differentiable Rendering SLAM}}}} \\

DDS-SLAM 
& 18.90 & 0.500 & 0.754
& 20.65 & 0.590 & 0.708
& 19.66 & 0.507 & 0.744
& 19.20 & 0.646 & 0.638 
& 19.60 & 0.561 & 0.711\\

EndoGSLAM
& 19.62 & 0.481 & 0.695
& 16.52 & 0.660 & 0.834
& 18.93 & 0.589 & 0.750
& 18.61 & 0.771 & 0.897
& 18.42 & 0.625 & 0.794 \\

Endo-2DTAM
& 18.10 & 0.455 & 0.661
& 17.95 & 0.611 & 0.733
& 17.93 & 0.551 & 0.721
& 16.86 & 0.581 & 0.711
& 17.71 & 0.550 & 0.707 \\

\midrule
\rowcolor{gray!10}
NRGS-SLAM (Ours)
& \textbf{24.62} & \textbf{0.654} & \textbf{0.595}
& \textbf{28.73} & \textbf{0.751} & \textbf{0.597}
& \textbf{26.45} & \textbf{0.696} & \textbf{0.633}
& \textbf{25.53} & \textbf{0.790} & \textbf{0.536}
& \textbf{26.33} & \textbf{0.723} & \textbf{0.590} \\

\bottomrule
\end{tabular}
\end{table*}
\begin{figure*}[th]
\centerline{\includegraphics[width=0.95\textwidth]{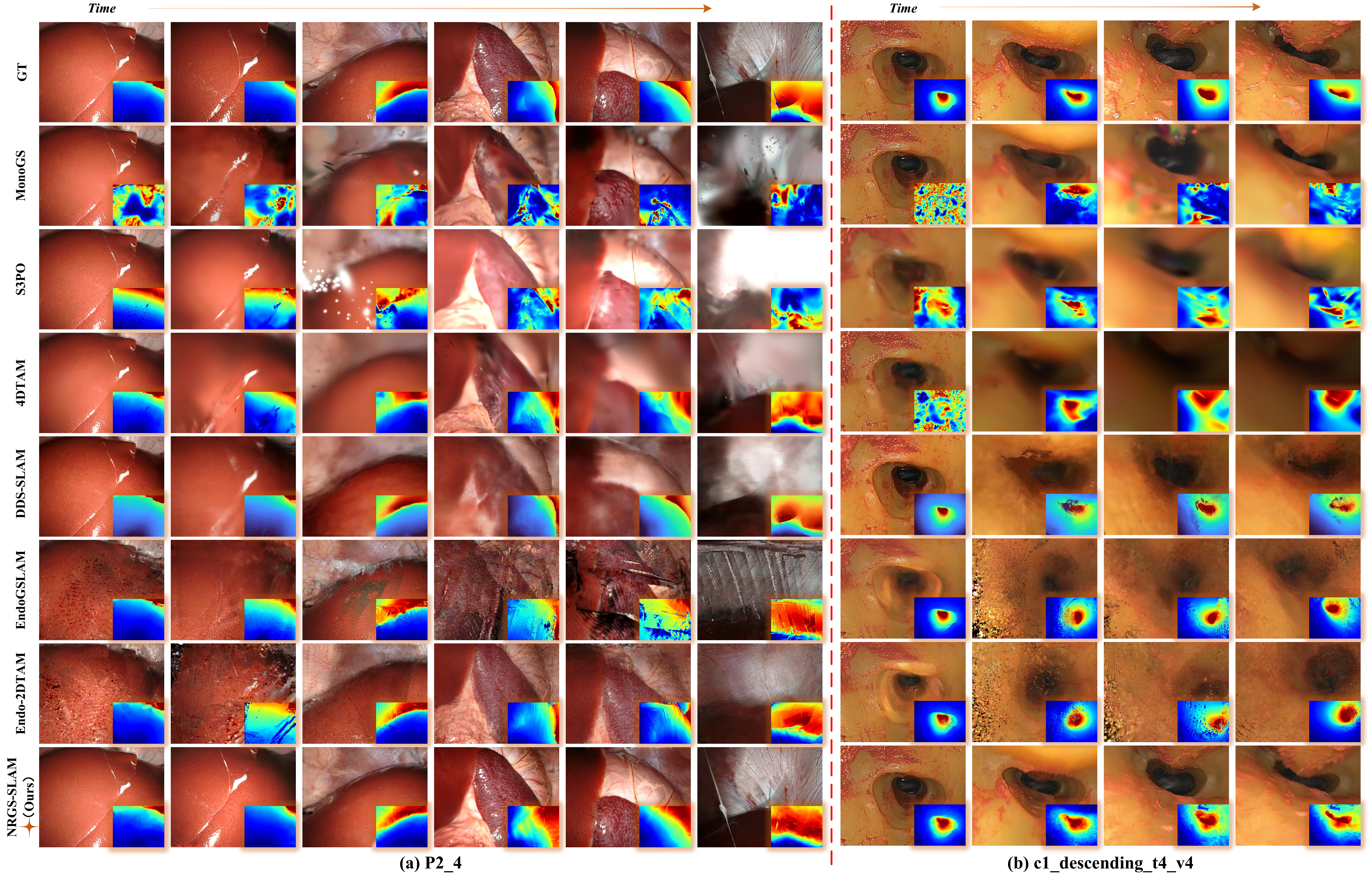}}
\caption{Qualitative comparison of rendering results across different methods at multiple time steps on StereoMIS and C3VDv2 datasets. For each method, the RGB rendering is shown with the corresponding depth map inset at the bottom right. On both datasets, the depth maps are included for reference only. For C3VDv2, the depth maps are inferred offline using \cite{xiao2025spatialtrackerv2}, while for StereoMIS, they are computed using the stereo model provided by the dataset \cite{hayoz2023learning}.}
    \label{fig:render}
\end{figure*}
\begin{figure}[t]
\centerline{\includegraphics[width=0.45\textwidth]{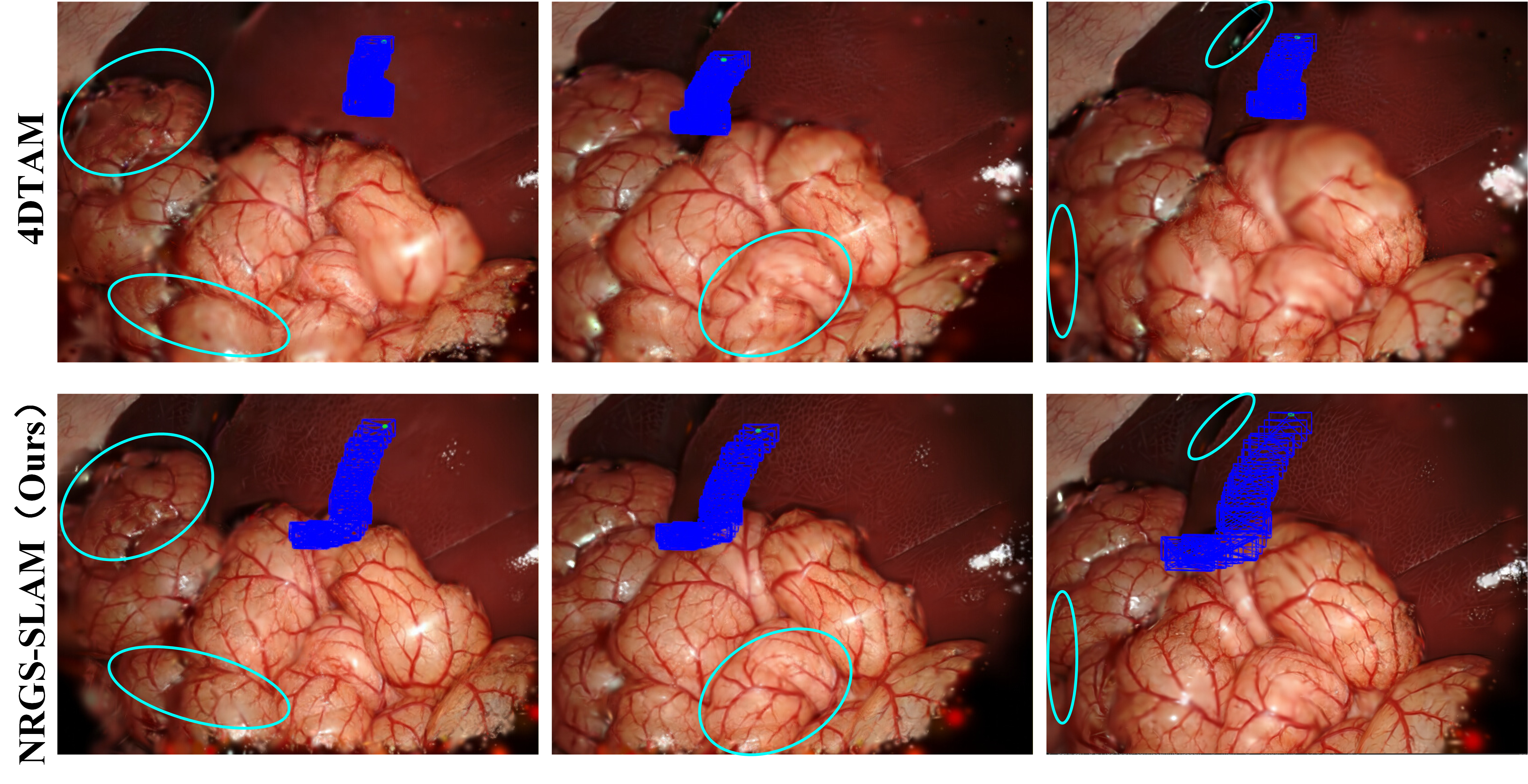}}
\caption{
Qualitative comparison of environment mapping results on Sequence P2-5 of the StereoMIS dataset.
The top row shows the results of 4DTAM and the bottom row shows the results of NRGS-SLAM.
The reconstructed camera trajectory is visualized in blue.
Cyan ellipses highlight regions for visual comparison.
NRGS-SLAM demonstrates improved structural completeness and higher-fidelity reconstruction with fewer artifacts compared to 4DTAM.
}
    \label{fig:3d vis}
\end{figure}

\textbf{Evaluation on C3VDv2 Dataset:}
Compared to the StereoMIS dataset, C3VDv2 contains shorter sequences (approximately 300 frames per sequence) but presents substantially greater challenges due to large-scale geometric deformations induced by external compression. The quantitative results are reported in Table~\ref{tab:camera_localization_c3vdv2}. Fig.~\ref{fig:traj c3vdv2} further illustrates the severe time-varying scene deformations and the corresponding camera trajectories estimated by different methods. Methods relying on rigid-scene assumptions, including general-purpose approaches such as MonoGS and S3PO as well as endoscopy-oriented methods such as EndoGSLAM and Endo-2DTAM, exhibit significantly higher localization errors. This performance degradation arises because the fundamental assumption of scene rigidity is violated. In contrast, non-rigid approaches, DDS-SLAM  and 4DTAM, achieve lower trajectory errors, underscoring the importance of deformation modeling. Our proposed NRGS-SLAM achieves the best overall performance with an average RMSE of 8.13 mm, indicating stable and accurate tracking even in the presence of substantial geometric and topological changes.

\subsubsection{Environment Mapping}

We compare different methods on multiple sequences from the StereoMIS, Hamlyn, and C3VDv2 datasets. Traditional SLAM systems (e.g., DefSLAM and NR-SLAM) do not support photorealistic rendering and are therefore excluded from this evaluation. Tables~\ref{tab:rendering_stereomis}, \ref{tab:rendering_hamlyn}, and \ref{tab:rendering_c3vdv2} report the quantitative rendering results, while Fig.~\ref{fig:render} and Fig.~\ref{fig:3d vis} provide qualitative comparisons. Overall, NRGS-SLAM achieves higher-quality deformable scene reconstruction than state-of-the-art methods across different deformation scenarios. Detailed analysis is provided below.

As shown in Fig.~\ref{fig:render}, on the StereoMIS dataset, GS-based methods originally designed for static scenes produce noticeable rendering artifacts, blurred textures, and geometric distortions as the camera moves and the scene undergoes non-rigid deformation.
Deformation-aware methods, DDS-SLAM and 4DTAM, alleviate these issues by explicitly modeling deformation.
Compared with all baselines, our proposed NRGS-SLAM produces visually coherent and photorealistic renderings while preserving fine texture details. The 3D reconstruction results at different time steps, illustrated in Fig.~\ref{fig:3d vis}, further support this observation. This advantage is also reflected in Tables~\ref{tab:rendering_stereomis}, where NRGS-SLAM achieves the best average rendering performance in both clip mode and full-sequence evaluation. Similar improvements are observed on the Hamlyn dataset, as shown in Tables~\ref{tab:rendering_hamlyn}.

As further illustrated in Fig.~\ref{fig:render}, the C3VDv2 dataset contains extremely large deformations, which pose significant challenges to all methods. EndoGSLAM and Endo-2DTAM fail to adapt to such non-rigid variations. As deformation accumulates, both geometry and appearance become severely distorted, resulting in incorrect surface structures and blurred textures.
Other rigid-scene methods, such as MonoGS and S3PO, exhibit similar issues as observed on StereoMIS, but the performance degradation is more severe. Compared with rigid baselines, DDS-SLAM and 4DTAM achieve higher PSNR and SSIM and lower LPIPS (Tables~\ref{tab:rendering_c3vdv2}). However, their renderings still contain texture blurring and local geometric inaccuracies due to limited deformation modeling capacity and accumulated optimization errors.
In contrast, as reported in Tables~\ref{tab:rendering_c3vdv2}, our proposed NRGS-SLAM achieves the best performance across all metrics on C3VDv2. The consistent improvements in PSNR, SSIM, and LPIPS indicate the effectiveness of the proposed non-rigid Gaussian representation and optimization strategy under extreme deformation conditions.

\subsection{Ablation Study}
In this section, we conduct a comprehensive ablation study to evaluate the effectiveness of the key designs within our modules on the StereoMIS (sequence P2\_5) and C3VDv2 (sequence c1\_sigmoid1\_t4\_v4) datasets.

\subsubsection{Measurement Preprocessing and Geometric Supervision}

We perform an ablation study on the Measurement Preprocessing module (Sec.~\ref{sec:mea}), jointly assessing the geometric supervision and robust optimization strategies. The quantitative results are reported in Table~\ref{tab:ablation_meas}.
Removing geometric priors entirely (\underline{\textit{Config.~A1}}) leads to a catastrophic performance drop across both datasets, with the RMSE surging by approximately 87\% on StereoMIS and 70\% on C3VDv2.
These results indicate that geometric priors are essential for constraining the ill-posed monocular non-rigid SLAM problem. 
Disabling the unified robust optimization strategy (\underline{\textit{Config.~A2}}), i.e., replacing it with a standard $\ell_2$ loss, increases the tracking error. This result highlights the importance of the annealed geometric weights and the IRLS-based reweighting in reducing the impact of unreliable priors and maintaining stable optimization.
Excluding the validity mask (\underline{\textit{Config.~A3}}) or the co-visibility mask (\underline{\textit{Config.~A4}}) consistently degrades tracking accuracy. 
This decline occurs because the validity mask is crucial for filtering unreliable measurements caused by specular reflections or low-illumination regions, while the co-visibility mask prevents incorrect constraints in unobserved areas.

\subsubsection{Deformable Tracking}

We investigate the effectiveness of the key design components in the Deformable Tracking module (Sec.~\ref{sec:tracking}), including Deformation-Aware Camera Tracking and Per-frame Deformation Estimation.

\begin{figure}[t]
\centerline{\includegraphics[width=0.5\textwidth]{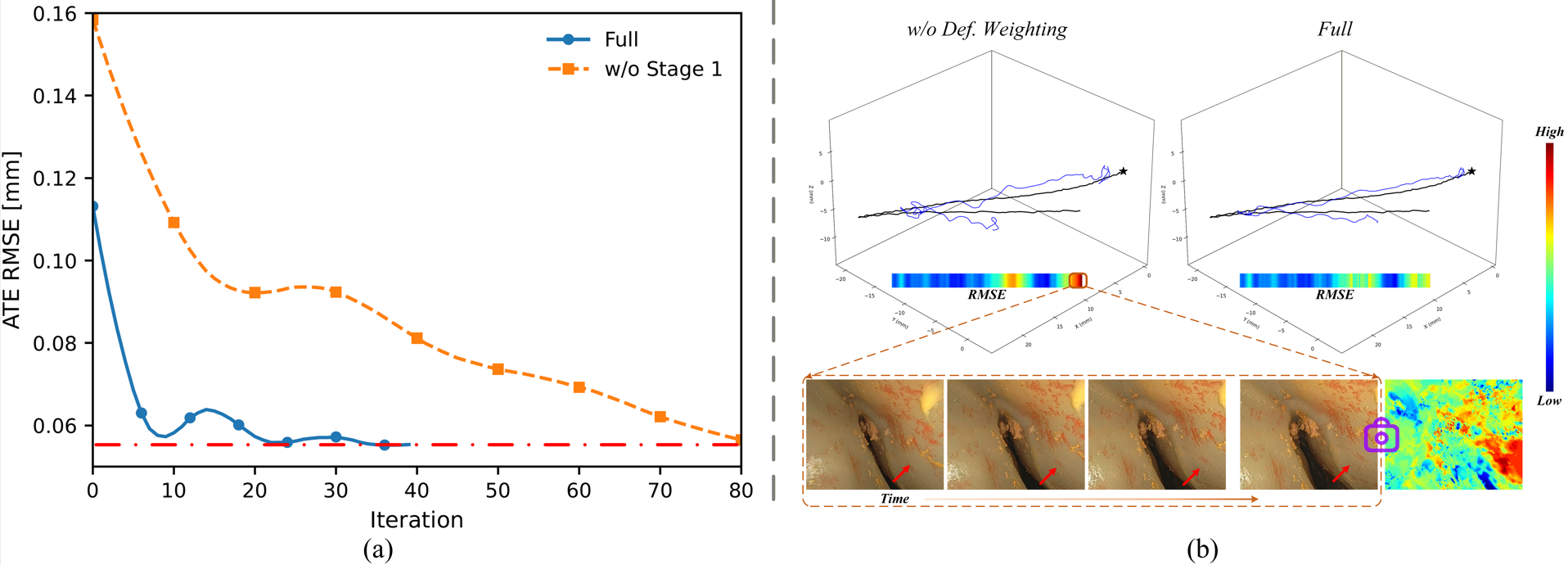}}
\caption{Ablation study of the Deformation-Aware Camera Tracking. 
(a) ATE RMSE versus optimization iterations for the full model and the variant without Stage 1. 
(b) Comparison between the variant without deformation-aware weighting (left) and the full model (right). 
The top row shows the estimated camera trajectories colored by RMSE. 
The bottom row presents the observed image sequence over a time interval with large tracking errors. 
The rightmost image shows the 2D deformation confidence map of the final frame.}
    \label{fig:tracking2 stage}
\end{figure}

\textbf{Deformation-Aware Camera Tracking.} 
As reported in Table~\ref{tab:ablation_tracking}, removing Stage~1 (PnP initialization, \underline{\textit{Config.~B1}}) increases the localization error on StereoMIS and C3VDv2.
Furthermore, as shown in Fig.~\ref{fig:tracking2 stage}(a), the full model exhibits faster error reduction in the early optimization stage and stabilizes earlier than the variant without Stage 1.
Moreover, removing the second stage refinement (\underline{\textit{Config.~B2}}) leads to a drastic degradation in localization accuracy, with RMSE surging by approximately 149\% on StereoMIS and 167\% on C3VDv2. This result indicates that PnP initialization alone is insufficient for reliable non-rigid SLAM.
Disabling the deformation-aware weighting (\underline{\textit{Config.~B3}}) also degrades performance, increasing the RMSE by approximately 46\% on StereoMIS and 69\% on C3VDv2. As shown in Fig.~\ref{fig:tracking2 stage}(b), the full model produces a more stable trajectory in regions with large tracking errors. The deformation-aware weighting assigns lower importance to regions undergoing strong non-rigid deformation, which reduces their adverse effect on camera pose optimization. 

\begin{table}[t]
\centering
\caption{Ablation Study on Measurement Preprocessing and Geometric Supervision.}
\label{tab:ablation_meas}
\scriptsize 
\setlength{\tabcolsep}{6pt} 
\renewcommand{\arraystretch}{1.1} 

\resizebox{\linewidth}{!}{
\begin{tabular}{lcccc}
\toprule
\multirow{2}{*}{{Configuration}} & \multicolumn{2}{c}{{StereoMIS}} & \multicolumn{2}{c}{{C3VDv2}} \\
\cmidrule(lr){2-3} \cmidrule(lr){4-5}
 & RMSE $\downarrow$ & S.D. $\downarrow$ & RMSE $\downarrow$ & S.D. $\downarrow$ \\
\midrule
\textit{A1. w/o Geo. Priors} & 13.85 & 6.42 & 13.55 & 6.65 \\
\textit{A2. w/o Robust Opt.} & 8.95 & 4.36 & 9.36 & 4.25 \\
\textit{A3. w/o Val. Mask}   & 7.92 & 3.32 & 8.42 & 3.91 \\
\textit{A4. w/o Co-vis. Mask}& 8.02 & 3.84 & 8.13 & 3.75 \\
\midrule
\textbf{Ours (Full)}        & \textbf{7.41} & \textbf{3.03} & \textbf{7.96} & \textbf{3.66} \\
\bottomrule
\end{tabular}
}
\end{table}
\begin{table}[t]
    \centering
    \caption{Ablation Study on Deformable Camera Tracking.}
    \label{tab:ablation_tracking}
    \setlength{\tabcolsep}{4pt}     \renewcommand{\arraystretch}{1.1}
    \begin{tabular}{lcccc}
    \toprule
    \multirow{2}{*}{Configuration} & \multicolumn{2}{c}{StereoMIS} & \multicolumn{2}{c}{C3VDv2} \\
    \cmidrule(lr){2-3} \cmidrule(lr){4-5}
     & RMSE $\downarrow$ & S.D. $\downarrow$ & RMSE $\downarrow$ & S.D. $\downarrow$ \\
    \midrule
    \textit{B1. w/o Stage 1 (PnP Init.)} & 8.75 & 3.94 & 8.84 & 4.12 \\
    \textit{B2. w/o Stage 2 (Refinement)} & 18.45 & 6.82 & 21.26 & 9.41 \\
    \textit{B3. w/o Def. Weighting} & 10.84 & 4.24 & 13.43 & 6.45 \\
    \midrule
    \textbf{Ours (Full Model)} & \textbf{7.41} & \textbf{3.03} & \textbf{7.96} & \textbf{3.66} \\
    \bottomrule
    \end{tabular}
\end{table}
\begin{table}[t]
    \centering
    \caption{{Ablation Study on Per-frame Deformation Estimation.} {FPS}: Frames Per Second.}
    \label{tab:ablation_deformation}
    \setlength{\tabcolsep}{2.5pt}
    \renewcommand{\arraystretch}{1.1}
    \begin{tabular}{lcccccc}
    \toprule
    \multirow{2}{*}{Configuration} & \multicolumn{3}{c}{StereoMIS} & \multicolumn{3}{c}{C3VDv2} \\
    \cmidrule(lr){2-4} \cmidrule(lr){5-7}
     & RMSE $\downarrow$ & S.D. $\downarrow$ & FPS $\uparrow$ & RMSE $\downarrow$ & S.D. $\downarrow$ & FPS $\uparrow$ \\
    \midrule
    \textit{C1. Full Param. Update} & {7.32} & {2.98} & 0.652 & {7.66} & {3.24} & 0.77 \\
    \midrule
    \textbf{Ours (Residual-based)} & 7.41 & 3.03 & 0.924 & 7.96 & 3.66 & 1.05 \\
    \bottomrule
    \end{tabular}
\end{table}
\begin{table}[t]
    \centering
    \caption{Ablation Study on Deformable Mapping.}
    \label{tab:ablation_mapping}
    \setlength{\tabcolsep}{2.5pt}
    \renewcommand{\arraystretch}{1.1}
    \begin{tabular}{lcccccc}
    \toprule
    \multirow{2}{*}{Configuration} & \multicolumn{3}{c}{StereoMIS} & \multicolumn{3}{c}{C3VDv2} \\
    \cmidrule(lr){2-4} \cmidrule(lr){5-7}
     & RMSE $\downarrow$ & S.D. $\downarrow$ & FPS $\uparrow$ & RMSE $\downarrow$ & S.D. $\downarrow$ & FPS $\uparrow$ \\
    \midrule
    \textit{D1. w/o Pose Opt.} & 8.04 & 3.56 & 0.96 & 8.24 & 3.71 & {1.09} \\
    \textit{D2. w/o Def. Prob. Est.} & 11.95 & 5.42 & {1.09} & 11.49 & 4.54 & {1.26} \\
    \textit{D3. w/o Dyn. Mgmt.} & 7.84 & 3.26 & 0.81 & 8.62 & 3.97 & 0.95 \\
    \textit{D4. w/o $w_d$-Gating} & 8.15 & 3.62 & 0.94 & 8.54 & 3.98 & 1.05 \\
    \midrule
    \textbf{Ours (Full Model)} & \textbf{7.41} & \textbf{3.03} & 0.92 & \textbf{7.96} & \textbf{3.66} & 1.05 \\
    \bottomrule
    \end{tabular}
\end{table}
\begin{figure}[t]
\centerline{\includegraphics[width=0.5\textwidth]{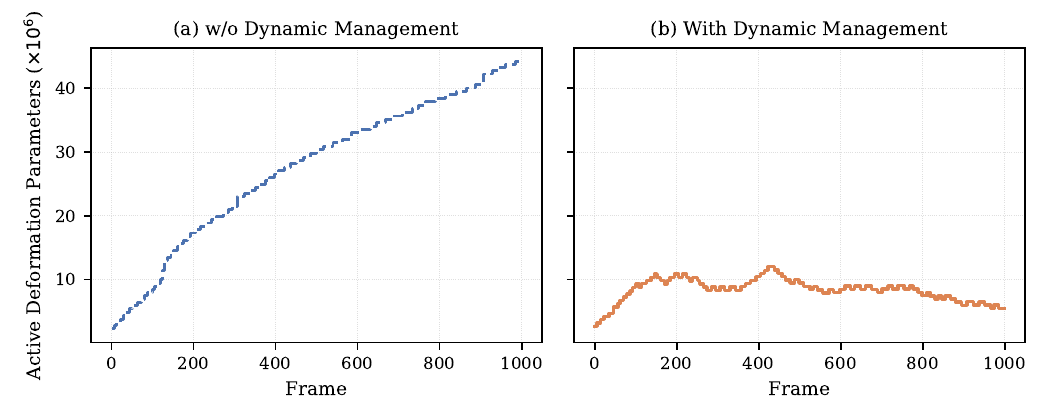}}
\caption{Impact of dynamic deformation management on the evolution of active deformation parameters. 
The number of active deformation parameters over frames for (a) without dynamic management and (b) with dynamic management. 
Without management, parameters grow steadily over time, whereas dynamic management maintains a bounded and stable model complexity.}
    \label{fig:def mana}
\end{figure}

\textbf{Per-frame Deformation Estimation.}
As shown in Table~\ref{tab:ablation_deformation}, although the full parameter update (\underline{\textit{Config.~C1}}) achieves slightly lower localization errors than our residual-based method on both datasets, it incurs a heavy computational burden. By contrast, the residual-based method preserves comparable tracking accuracy while increasing the frame rate by approximately 42\% on StereoMIS and 36\% on C3VDv2.

\subsubsection{Deformable Mapping}
We conduct an ablation study on the key components of Deformable Mapping (Sec.~\ref{sec:mapping}) and the related designs in Sec.~\ref{sec:map}, given their close coupling with the deformation-aware 3D Gaussian map.

\textbf{Impact of Keyframe Poses Optimization.} 
As shown in Table~\ref{tab:ablation_mapping} (\underline{\textit{Config.~D1}}), removing the sliding-window pose optimization increases the RMSE from 7.41 to 8.04 on StereoMIS and from 7.96 to 8.24 on C3VDv2. This module performs several refinement iterations to mitigate accumulated drift, providing reliable pose initialization for subsequent map expansion and deformation probability estimation.

\textbf{Impact of Deformation Probability Estimation.}
As shown in Table~\ref{tab:ablation_mapping} (\underline{\textit{Config.~D2}}), removing the Deformation Probability Estimation module significantly degrades tracking accuracy. Without this guidance, the deformation probability cannot be properly learned, which compromises the disentanglement between camera ego-motion and scene deformation during optimization. As a result, the RMSE increases by approximately 61\% on StereoMIS and 44\% on C3VDv2.

\textbf{Impact of Dynamic Deformation Field Management.}
Disabling dynamic deformation management (\underline{\textit{Config.~D3}}) degrades both accuracy and efficiency, as redundant deformation parameters accumulate over time, increasing optimization complexity and reducing numerical stability. This parameter growth slows down the system and biases the deformation updates, resulting in higher localization errors. As illustrated in Fig.~\ref{fig:def mana}, the management strategy constrains parameter growth and maintains stable computational cost over long sequences.

\textbf{Impact of $w_{d,i}$-Gating.}
As shown in Table~\ref{tab:ablation_mapping} (\underline{\textit{Config.~D4}}), removing the $w_{d,i}$ gating increases the RMSE to 8.15 on StereoMIS and 8.54 on C3VDv2. Without this modulation, deformation is applied uniformly to all Gaussians regardless of rigidity, leading to erroneous warping of static regions.

\begin{table}[t]
\centering
\caption{Runtime Analysis.}
\setlength{\tabcolsep}{3pt}
\label{tab:runtime}
\begin{tabular}{lccccc}
\hline
{Dataset-Scene} & \begin{tabular}[c]{@{}c@{}}{Frame}\\{Number}\end{tabular} & 
\begin{tabular}[c]{@{}c@{}}{Total}\\{Runtime}\end{tabular} & 
\begin{tabular}[c]{@{}c@{}}{Tracking}\\{/Frame}\end{tabular} & 
{FPS} \\
\hline
StereoMIS-P2\_4 & 4281 & 4896s & 232ms & 0.874 \\
C3VDv2-C1\_descending\_t4\_v4  &361 & 383s  & 209ms & 0.943 \\
\hline
\end{tabular}
\end{table}
\begin{figure}[t]
\centerline{\includegraphics[width=0.5\textwidth]{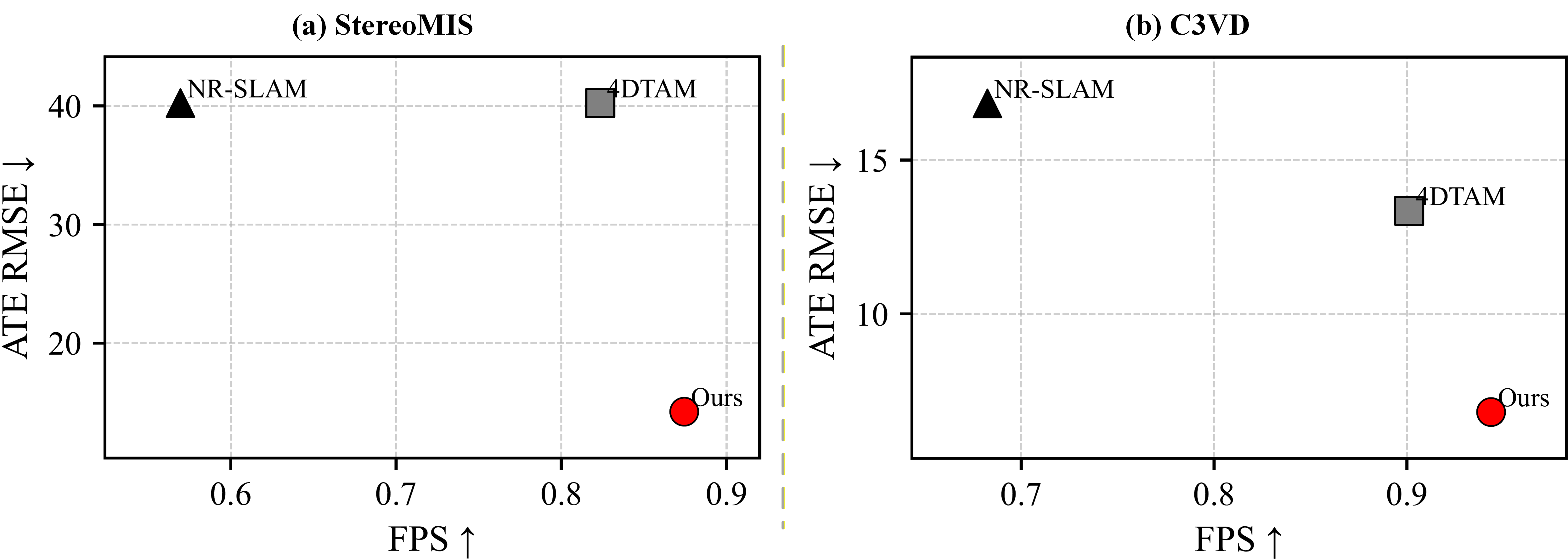}}
\caption{Speed--accuracy trade-off on two datasets. Higher FPS indicates faster processing, while lower ATE RMSE indicates better accuracy.}
    \label{fig:runtime}
\end{figure}

\subsection{Runtime Analysis}

Table~\ref{tab:runtime} reports the runtime statistics on two representative sequences from StereoMIS and C3VDv2, where our method achieves 0.874 FPS on P2\_4 and 0.943 FPS on C1\_descending\_t4\_v4.
Although the current implementation does not reach real-time performance, the computational cost is reasonable given the inherent complexity of non-rigid SLAM, which requires jointly estimating camera motion, scene geometry, and dense deformation fields. 
As shown in Fig.~\ref{fig:runtime}, both traditional and GS-based non-rigid SLAM methods exhibit increased computational cost due to deformation modeling. However, our method achieves a superior balance, delivering significantly lower trajectory error while maintaining a frame rate comparable to, or even slightly higher than, competing state-of-the-art deformable systems.

\section{Limitations}

Our proposed NRGS-SLAM is built on a conceptually straightforward yet technically challenging premise: distinguishing between quasi-rigid and deformable regions within endoscopic scenes. By prioritizing visual observations from quasi-rigid regions for camera tracking, the system achieves accurate and robust pose estimation. These pose estimates then support high-fidelity deformable scene reconstruction using 3D Gaussian Splatting, forming a mutually reinforcing optimization process. Experimental and ablation results demonstrate the effectiveness of this design.

Beyond these immediate results, the proposed framework has potential for downstream applications such as preoperative-to-intraoperative registration. A core challenge in this domain is aligning the deformed intraoperative scene reconstructed from endoscopy with a static preoperative 3D model. By identifying quasi-rigid regions during surgery, the method provides reliable reference structures for alignment and can potentially serve as a robust initialization for registration. Additionally, although the current system does not operate in real time, its processing speed is sufficient for applications such as high-fidelity postoperative analysis and surgical training playback, where reconstruction accuracy takes precedence over immediate interactivity.

However, several limitations remain for practical deployment. First, the system does not yet achieve real-time performance. This is primarily attributed to the deformation field representation. Similar to many existing methods that explicitly model Gaussian deformation \cite{yang2024deform3dgs,shan2025deformable}, the approach assigns deformation parameters to each Gaussian primitive individually, resulting in a high-dimensional optimization problem that increases computational cost. In practice, deformation behavior is often locally consistent across a surface, as leveraged by NR-SLAM\cite{rodriguez2024nr}. Future work may explore modeling deformation at the surface level rather than at the primitive level, which would reduce parameter dimensionality and improve runtime efficiency.

Another limitation lies in the deformation probability estimation module. While it provides an effective posterior for deformation learning, it introduces additional computational overhead. The module alternates with the optimization process during global bundle adjustment and requires two additional rendering passes (under rigid and deformable hypotheses) at each step. As shown in Table~\ref{tab:ablation_mapping} (Config.~D2), removing this module increases the frame rate from 0.92 to 1.09 FPS on StereoMIS, and from 1.05 to 1.26 FPS on C3VDv2. Furthermore, the module primarily relies on RGB observations to infer the posterior. In endoscopic environments, visual cues are frequently degraded by specular reflections, illumination variation, and weak tissue texture, making it challenging to distinguish rigid motion from non-rigid deformation. Future research will investigate stronger regularization strategies and the integration of complementary geometric or temporal cues to improve robustness. Leveraging foundation models may further alleviate the inherent ill-posedness of deformable SLAM.

Finally, multi-modal sensor fusion represents an important direction toward clinical deployment. By integrating the visual reconstruction framework with complementary sensing modalities, the system can exploit cross-modal constraints to enhance robustness and accuracy. For example, Fiber Bragg Grating (FBG) sensors measure the shape of the surgical robot. Although FBG measurements are subject to their own sources of error, they are independent of visual tissue deformation and can potentially assist in distinguishing rigid from deformable regions. Conversely, visual observations from the endoscope can constrain camera pose estimation and help compensate for accumulated drift in FBG-based shape reconstruction.

\section{Conclusion}
This paper presents a monocular deformable SLAM system for deformable endoscopic scenes, enabling accurate camera pose estimation and high-quality photorealistic reconstruction. The core innovation is a deformation-aware 3D Gaussian map that assigns a learnable deformation probability to each primitive. Through a Bayesian self-supervision strategy, the system separates rigid camera motion from non-rigid tissue deformation without requiring external deformation annotations.  
Based on this representation, we develop a deformable tracking module that provides reliable pose estimation together with efficient per-frame deformation updates. We also introduce a deformable mapping module that incrementally expands and refines the scene representation while balancing modeling capacity and computational cost. In addition, geometric priors are incorporated through a unified robust loss formulation to alleviate the intrinsic ill-posedness of monocular non-rigid SLAM.  
Extensive experiments on public endoscopic datasets show that our method achieves competitive performance in both trajectory estimation accuracy and photorealistic scene reconstruction compared with existing state-of-the-art approaches. Ablation studies further confirm the contribution of each system component.

\bibliographystyle{IEEEtran}
\bibliography{root}

\clearpage

\end{document}